\pdfoutput=1

\documentclass[11pt]{article}
\usepackage[final]{acl}
\usepackage{times}
\usepackage{latexsym}
\usepackage[T1]{fontenc}
\usepackage[utf8]{inputenc}

\usepackage{microtype}

\usepackage{inconsolata}
\usepackage{tcolorbox}
\tcbuselibrary{skins,breakable}
\usepackage{amsmath}
\usepackage{bm}
\usepackage{booktabs}
\usepackage{graphicx}
\usepackage[table]{xcolor}
\usepackage{enumitem}
\setlist[]{left=0mm,noitemsep,topsep=0mm}
\usepackage{hyperref}
\usepackage{cleveref}
\usepackage{subcaption}
\usepackage{xspace}
\usepackage{multirow}

\usepackage{tikz}
\usepackage{kotex}        
\usepackage{fontawesome5}

\usetikzlibrary{positioning, calc}
\definecolor{segcolor}{HTML}{4A7EAB}    
\definecolor{doccolor}{HTML}{4A8E50}  
\definecolor{mixcolor}{HTML}{A04545}    

\newcommand{\seg}{\textcolor{segcolor}{\textbf{\textsf{\small SEG}}}\xspace}
\newcommand{\doc}{\textcolor{doccolor}{\textbf{\textsf{\small DOC}}}\xspace}
\newcommand{\docshuff}{\textcolor{mixcolor}{\textbf{\textsf{\small MIX}}}\xspace}

\newcommand{\xcomet}{x\textsc{comet-xxl}\xspace}            
\newcommand{\xcometkiwi}{\textsc{CometKiwi-xxl}\xspace}    
\newcommand{\metricx}{\textsc{metricX-24-xxl}\xspace}          
\newcommand{\metricxqe}{\textsc{metricX-24-qe-xxl}\xspace}     
\newcommand{\chrf}{chr\textsc{F}\xspace}                
\newcommand{\dbleu}{d-\textsc{bleu}\xspace}
\newcommand{\doccomet}{doc-\textsc{comet}\xspace}
\newcommand{\slide}{\textsc{slide}\xspace}
\newcommand{\segale}{\textsc{segale}\xspace}
\newcommand{\falcon}{\textsc{falcon}\xspace}
\newcommand{\gemba}{\textsc{gemba v2}\xspace}

\newcommand{\mGpt}{\textsc{GPT-4.1}\xspace}
\newcommand{\mGemini}{\textsc{Gemini-2.5-Pro}\xspace}
\newcommand{\mRef}{\faUser{}\textsc{REF}\xspace}
\newcommand{\mClaude}{\textsc{Claude-4}\xspace}
\newcommand{\mCommandA}{\textsc{CommandA-MT}\xspace}
\newcommand{\mGemtrans}{\textsc{GemTrans}\xspace}
\newcommand{\mIrb}{\textsc{irb-mt}\xspace}
\newcommand{\mYolu}{\textsc{yolu}\xspace}
\newcommand{\mLanqio}{\textsc{Laniqo}\xspace}
\newcommand{\mWenyiil}{\textsc{Wenyiil}\xspace}

\newcommand{\faHF}{\raisebox{-0.15ex}{\includegraphics[height=1em]{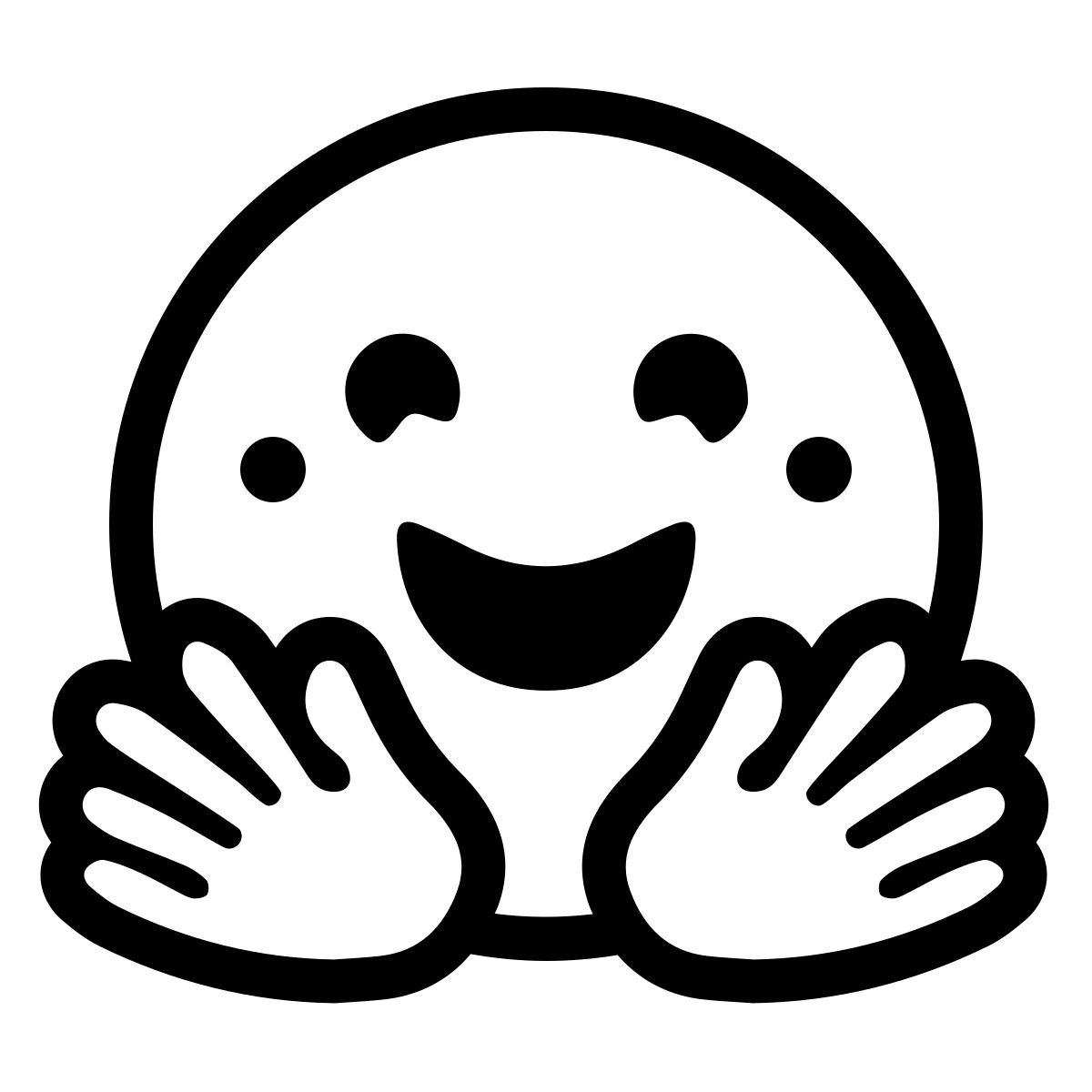}}}
\Crefname{appendix}{Appendix}{Appendices}

\newcounter{cond}
\renewcommand{\thecond}{\arabic{cond}}

\crefformat{cond}{#2H#1#3}
\Crefformat{cond}{#2H#1#3}
\crefrangeformat{cond}{H#3#1#4 to~H#5#2#6}
\Crefrangeformat{cond}{H#3#1#4 to~H#5#2#6}

\newtcolorbox[use counter=cond]{condbox}[2][]{
    enhanced, 
    colback=gray!5,
    colframe=gray!20,
    coltitle=black,
    fonttitle=\bfseries,
    title={H\thecond: #2},
    label type=cond,
    label={#1},
    left=3pt, 
    right=3pt, 
    top=2pt, 
    bottom=2pt,
    sharp corners,
    boxrule=0pt,
    underlay={
        \begin{tcbclipframe}
            \fill[black!70] (frame.south west) rectangle ([xshift=3pt]frame.north west);
        \end{tcbclipframe}
    },
    before skip=6pt,
    after skip=6pt
}

\newcommand{\circled}[1]{%
    \tikz[baseline=(char.base)]{
        \node[shape=circle, draw, inner sep=1pt, 
              font=\small\bfseries, line width=0.6pt, fill=gray!20] 
        (char) {#1};
    }%
}
\newcommand{\distbar}[3]{%
\begin{tikzpicture}[baseline=-0.5ex]
  \fill[blue!60]  (0,0)                 rectangle ({#1*0.025},0.22);
  \fill[red!50]   ({#1*0.025},0)        rectangle ({(#1+#2)*0.025},0.22);
  \fill[gray!40]  ({(#1+#2)*0.025},0)   rectangle ({(#1+#2+#3)*0.025},0.22);
\end{tikzpicture}%
}
\definecolor{LLMcolor}{HTML}{EE9B51}
\makeatletter
\newcommand\blfootnote[1]{%
  \begingroup
    \renewcommand\thefootnote{}%
    \let\orig@makefntext\@makefntext
    \def\@makefntext##1{\noindent##1}%
    \footnotetext{#1}%
    \addtocounter{footnote}{0}%
    \let\@makefntext\orig@makefntext
  \endgroup
}
\makeatother

\title{The Blindness of Document-Level Translation Evaluation}

\author{
  Ahrii Kim\textsuperscript{1} \quad
  Vilém Zouhar\textsuperscript{4} \quad
  Chanjun Park\textsuperscript{2} \quad
  Seong-heum Kim\textsuperscript{1,3 \thanks{Corresponding author.}} \\
  \textsuperscript{1}AI-Bio Convergence Research Inst. \quad
  \textsuperscript{2}School of Software \\
  \textsuperscript{3}Dept. of Intelligent Semiconductors \quad
  \textsuperscript{4}{ETH Zurich}\\
  Soongsil University \\
  \texttt{\{ahriikim,chanjun.park,seongheum\}@ssu.ac.kr} \quad \texttt{vzouhar@ethz.ch} \\
}

\begin{document}
\maketitle

\blfootnote{\faHF{}~\url{https://huggingface.co/datasets/trotacodigos/esa-counterfactual-enko}
\\\faGithub{}~\url{https://github.com/trotacodigos/esa-counterfactual}}

\begin{abstract}
Document-level machine translation (MT) evaluation extends segment-level protocols by presenting full documents to annotators, on the assumption that such presentation elicits document-level judgments. We test this assumption with a counterfactual condition (\docshuff) in which each document combines segments drawn from different systems, preserving document-level presentation while breaking cross-segment consistency. Across 18{,}420 expert English→Korean annotations and 14 automatic metrics, scores, system rankings, and error annotations are statistically equivalent between coherent and incoherent documents. Perception does not explain this: shown matched passages, raters identify the coherent one as the work of a single translator in 87.3\% of trials. Document presentation does change how annotators work, but that change does not reach the recorded output. What is blind is the protocol, not the annotator. The concern is not that scores fall short, but that the resources invested in document-level systems, metrics, and annotation may not be measuring what they are intended to measure.
\end{abstract}

\section{Introduction}
Document-level machine translation (MT) has been described as the missing piece for reaching human-level translation performance~\citep{kocmi-etal-2025-findings}. Early claims of human parity, on closer inspection, reflected the mismatch between segment-level and document-level translation~\citep{laubli-etal-2018-machine, popel2020transforming, toral-2020-reassessing}. Yet a series of puzzling patterns has accumulated in document-level MT evaluation: non-robust system rankings \citep{freitag-etal-2024-llms, kocmi-etal-2024-findings, kocmi-etal-2025-findings}, human scores clustered near the top of the scale~\citep{kocmi-etal-2024-findings}, surface metrics like \chrf~\citep{popovic-2015-chrf} performing comparably to large language model (LLM)-based ones~\citep{lavie-etal-2025-findings}, and literary text emerging as the easiest domain~\citep{kocmi-etal-2025-findings}. These anomalies persist in part because the validity of current document-level evaluation protocols has not been thoroughly confirmed.

\begin{figure}
    \centering
    \includegraphics[width=1\linewidth]{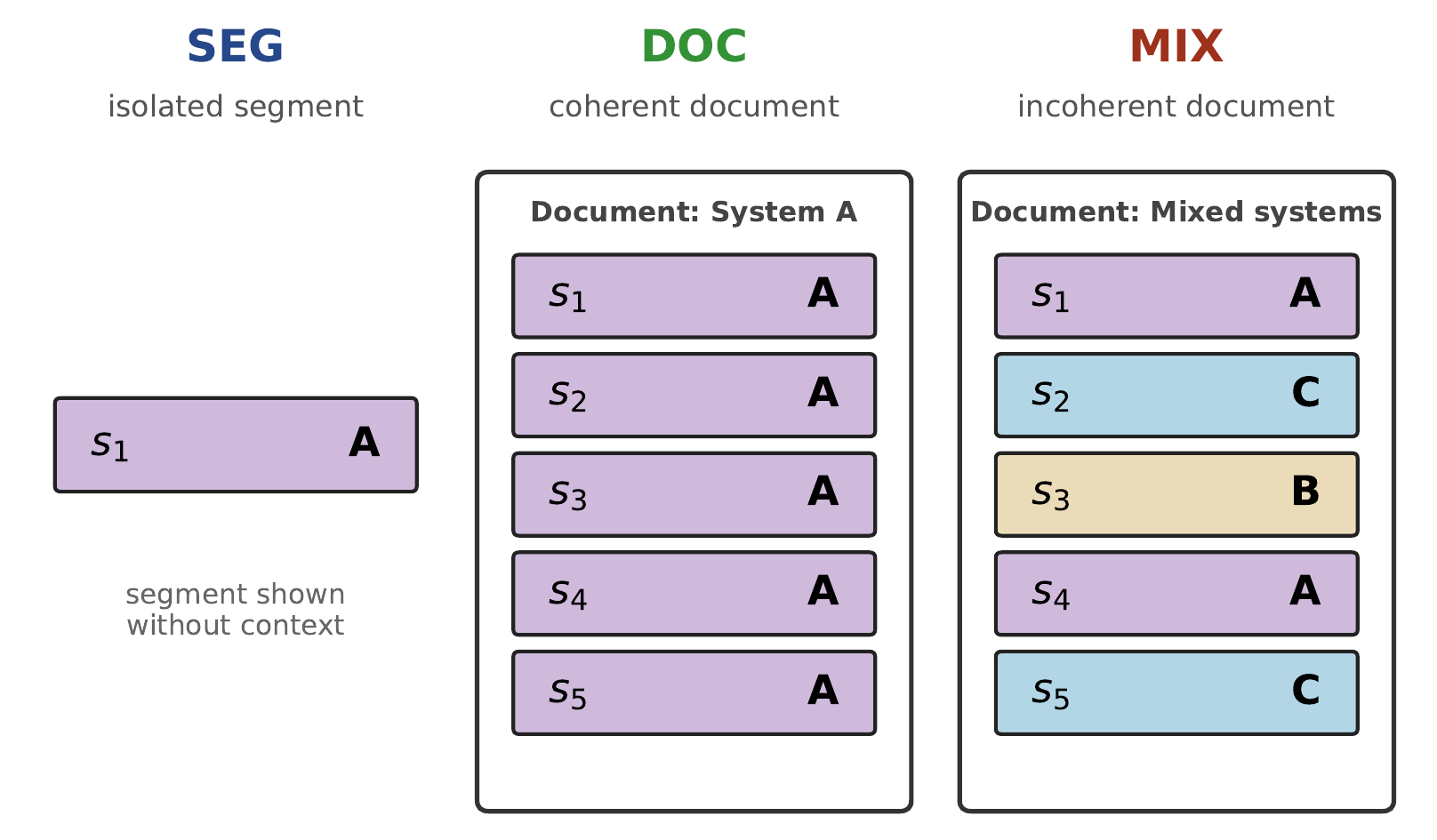}
    \caption{The three evaluation conditions. \seg presents a segment in isolation, without surrounding context. \doc presents a coherent document produced by a single system (here, System A). \docshuff presents a document of the same form but assembled from segments of different systems (A, B, C), preserving document-level layout while disrupting cross-segment coherence. 
    }
    \label{fig:main}
\end{figure}

Cognitive science offers a useful lens. Human evaluators can fail to notice quality issues through \textit{content-driven blindness}, where familiar or expected content reduces scrutiny~\citep{nick1998}, or through \textit{presentation-driven blindness}, where attention is shaped by how information is structured or displayed~\citep{MackRock1998InattentionalBlindness, SimonsChabris1999Gorillas, TREISMAN198097}. In MT evaluation, both raise the possibility that document-level cues go unregistered. The distinction we will need throughout is between document-level \emph{presentation}, which current protocols do provide, and document-level \emph{judgment}, which they assume follows from it. A cue can be perceived and still go unregistered if the instrument offers nowhere to put it. Our results place the failure at that second step: what is blind is the protocol, not the annotator.

To probe this possibility experimentally, we introduce a controlled counterfactual condition (\docshuff) in which each document combines segments (sentences or paragraphs) from multiple MT systems, preserving document-level presentation while intentionally removing discourse coherence (\Cref{fig:main}). We compare evaluation behavior and quality on \docshuff against coherent documents (\doc) and isolated segments (\seg) under both human annotation with Error Span Annotation (ESA;~\citealp{kocmi-etal-2024-error}) and 14 automatic metrics. We invest in depth over breadth, collecting 18,420 expert annotations on English$\rightarrow$Korean under a within-annotator paired design.
Korean's discourse-sensitive features offer a stress test for ESA, and establishing equivalence requires statistical power that thinner multi-language designs cannot achieve at comparable cost. A blind positive control confirms that the manipulation lands: shown a \doc and a \docshuff passage matched on segment quality, raters identify \doc as the work of a single consistent translator in $87.3\%$ of trials. We find that:

\begin{enumerate}
    \item Human ESA annotation yields statistically equivalent scores, near-identical system rankings, and largely indistinguishable error annotations between coherent (\doc) and incoherent (\docshuff) documents.

    \item All 14 automatic metrics tested produce the same system rankings on \doc and \docshuff. Only human evaluation shows slight divergence.
    
    \item The equivalence is not explained by annotators ignoring context. Score recoverability, agreement, and time-on-task all indicate that document presentation \emph{does} change annotator behavior, and ESA simply provides no channel in which that change can register.
\end{enumerate}

As context-aware modeling matures, the construct-validity of document-level evaluation becomes more urgent.
Our results suggest that, under current evaluation protocols, much of what is reported as document-level progress may reflect segment-level signal aggregated at the document level.
The concern is not that a measure has been corrupted by being targeted, but that the measure and the target are misaligned \citep{goodhart1984monetary}.
We take this as a reason to revisit discourse-level evaluation before it further misinforms the efforts built on top of it.

\section{Related Work}
\label{sec:related_work}

\subsection{Human Evaluation of Document MT}

Human evaluation has evolved through several protocols, all sharing a segment-level core. Direct Assessment (DA;~\citealp{graham-etal-2013-continuous}) aggregates segment-level scores into document-level averages~\citep{bojar-etal-2017-findings}. Early holistic document-level scoring suffered from low inter-annotator agreement \citep[IAA;][]{laubli-etal-2018-machine, castilho-2020-page}, and the field subsequently converged on segment-level annotation with document context~\citep{kocmi-etal-2022-findings}. Within this paradigm, Multidimensional Quality Metrics (MQM;~\citealp{freitag-etal-2021-experts}), ESA~\citep{kocmi-etal-2024-error}, and RATE~\citep{popov-etal-2025-refined} have refined error taxonomies and annotation interfaces, while recent work has expanded the evaluation criteria to include document-level error categories~\citep{kim-2025-context, song-etal-2025-enhancing}.
Still, the existing approaches remain segment-centric: document context is provided as a cue for segment scoring, with no direct evidence that the resulting annotations capture document-level quality.

\subsection{Automatic Evaluation of Document MT}

\paragraph{Context-Extended Metrics.}
One line of work extends existing metrics with broader context. \dbleu~\citep{liu-etal-2020-multilingual-denoising} and \doccomet~\citep{vernikos-etal-2022-embarrassingly} operate on full documents, while training-free alternatives prepend preceding context to pretrained metrics~\citep{vernikos-etal-2022-embarrassingly} or aggregate adjacent sentences via sliding windows~\citep[\slide;][]{raunak-etal-2023-evaluating}. \segale~\citep{wang-etal-2025-extending} scales this approach to book-length documents through adaptive sentence alignment. Local windows, however, may not capture phenomena requiring global coherence, and for full-document inputs it remains unclear whether the metrics encode document-level information or merely exploit extended surface context.

\paragraph{Discourse-Targeted Metrics.}
A second line targets specific phenomena: pronoun resolution (\textsc{autoprf},~\citealp{hardmeier-federico-2010-modelling}; \textsc{apt},~\citealp{miculicich-werlen-popescu-belis-2017-validation}; \textsc{protest},~\citealp{guillou-hardmeier-2018-automatic}), lexical choice (\textsc{atec},~\citealp{wong-kit-2010-parameter}), and multi-phenomenon evaluation covering tense, named entities, and terminology~\citep{sun-etal-2022-rethinking, jiang-etal-2022-blonde, wang-etal-2023-document-level}. These metrics offer fine-grained coverage but rely on surface-level string matching, have been validated only on their target languages, and do not address cross-sentential phenomena such as narrative coherence. \textsc{muda} \citep{fernandes-etal-2023-translation} instead releases taggers that locate context-dependent phenomena in arbitrary datasets, offering a route to quantifying per phenomenon how much inconsistency our counterfactual introduces.

\paragraph{LLM-as-a-judge.}
LLM-as-judge approaches have recently been applied to discourse-level evaluation, prompting LLMs to score document-level phenomena such as coherence~\citep{Kim_2025-falcon, sun-etal-2025-fine} or specific properties such as coreference and formality~\citep{mohammed-etal-2026-unlocking}. These methods rely on loosely specified criteria that have produced low IAA and have been reported to perform inconsistently on discourse-level phenomena~\citep{xiao-etal-2011-document, sun-etal-2025-fine, kim-2025-context}.

\vspace{1em}
Perturbation has been used to probe context sensitivity on the modeling side. \citet{mohammed-niculae-2025-context} perturb and randomize the document context supplied to nine LLMs and two encoder–decoder baselines, and find that gains in document-level translation quality do not carry over to pronoun translation. The logic is the one we adopt: a system that genuinely uses context should react when that context is corrupted. We apply it one level up, to the instrument rather than the model. Across all lines of work, an assumption has gone untested: that providing document-level input, whether as annotator context or as extended metric input, is sufficient for evaluation to operate at the document level. We test this assumption experimentally.

\section{Methods}
\label{sec:method}

\paragraph{The ESA protocol.}
Given a document $d$ translated by system $s \in S$ and split into $n$ segments, ESA elicits one score per segment while presenting the full document as context (\doc). Let $e_{s,k}$ denote the score assigned to segment $k$ under system $s$. The document-level score is the average of segment scores:

\begin{equation}
    E_s(d) = \frac{1}{n}\sum_{k=1}^n e_{s,k}
\end{equation}

\noindent The design intent of ESA is that document-level presentation allows annotators to register information not available from any segment in isolation (\seg). To make this intent explicit, we decompose each segment score as

\begin{equation}
    e_{s,k} = e_{s,k}^{\text{seg}} + e_{s,k}^{\text{ctx}},
\end{equation}

\noindent where $e_{s,k}^{\text{seg}}$ is the score the segment would receive if judged in isolation, and $e_{s,k}^{\text{ctx}}$ is the document-level cues induced by access to cross-segment context. ESA thereby assumes that document-level effects are captured through $e_{s,k}^{\text{ctx}}$, provided that the full document is visible and judgments are recorded at the segment level.

\paragraph{The counterfactual ESA protocol.}
We introduce a counterfactual condition (\docshuff) that preserves the interface constraints of ESA, namely document-level display with per-segment judgments, while removing genuine document-level coherence by construction. A mixed document $d^{\text{mix}}$ is assembled from segments of multiple systems, retaining each segment's within-document position but disrupting cross-segment coherence. Annotators provide segment scores $e_{s,k}^{\text{mix}}$ under the standard ESA interface. The counterfactual document score is defined analogously:

\begin{equation}
    E_s^{\text{mix}}(d) = \frac{1}{n}\sum_{k=1}^n e_{s,k}^{\text{mix}}
\end{equation}

\noindent Using the same additive decomposition,

\begin{equation}
    e_{s,k}^{\text{mix}} = e_{s,k}^{\text{seg}} + e_{s,k}^{\text{ctx,mix}}
\end{equation}

\noindent where $e_{s,k}^{\text{ctx,mix}}$ captures any context-driven cues induced by the mixed presentation. If ESA registers cross-segment coherence as intended, the presence of incoherent context under \docshuff should drive $e_{s,k}^{\text{ctx,mix}}$ in a direction distinguishable from $e_{s,k}^{\text{ctx}}$, and we know that $e_{s,k}^{\text{ctx,mix}} \neq 0$.

\paragraph{Context cues.}
ESA produces three outputs per segment that can carry $e_{s,k}^{\text{ctx,mix}}$: \circled{1} a numeric score on a 100-point scale; \circled{2} error span annotations, in which annotators highlight problematic spans; and \circled{3} a severity label (Major/Minor) on each span. If $e_{s,k}^{\text{ctx,mix}}$ differs from $e_{s,k}^{\text{ctx}}$, we expect \docshuff to depart from \doc along these signals. Scores should be lower under \docshuff, errors should be more frequent and more severe, and highlighted spans would concentrate at segment boundaries, where transitions between adjacent sentences are most visible.

\section{Experiment}

\subsection{Hypotheses}
\label{subsec:method}
We formalize three predictions entailed by ESA's design, if it registers cross-segment coherence as intended, for the contrast between \doc and \docshuff using three signals introduced above (\circled{1}, \circled{2}, and \circled{3}). For condition $c$, let $\mu_c$ denote the mean segment-level score, $r_c$ the induced system ranking, and $\eta_c$ the mean error count per annotation. $\tau_{a,b} = \tau(r_a, r_b)$ denotes Kendall's $\tau$ between rankings under conditions $a$ and $b$. We pre-specify equivalence margins $\epsilon$ and $\epsilon_{\mathrm{err}}$ for scores and error counts, and a lower bound $\tau_{\min}$ for ranking agreement. Subscripts $\text{\small DOC}$ and $\text{\small MIX}$ refer to \doc and \docshuff.

\begin{condbox}[h1]{Score Equivalence}
$H_0: |\mu_{\text{\tiny DOC}} - \mu_{\text{\tiny MIX}}| \geq \epsilon$
\\
$H_1: |\mu_{\text{\tiny DOC}} - \mu_{\text{\tiny MIX}}| < \epsilon$
\end{condbox}

\begin{condbox}[h2]{Ranking Agreement}
$H_0: \tau_{\text{\tiny DOC}, \text{\tiny MIX}} < \tau_{\min}$
\\
$H_1: \tau_{\text{\tiny DOC}, \text{\tiny MIX}} \geq \tau_{\min}$
\end{condbox}

\begin{condbox}[h3]{Error Equivalence}
$H_0: |\eta_{\text{\tiny MIX}} - \eta_{\text{\tiny DOC}}| \geq \epsilon_{\mathrm{err}}$
\\
$H_1: |\eta_{\text{\tiny MIX}} - \eta_{\text{\tiny DOC}}| < \epsilon_{\mathrm{err}}$
\end{condbox}

\vspace{1em}
\noindent
Our primary comparison is between \doc and \docshuff. We additionally compare \seg with \doc to check whether the document-level presentation format affects evaluation outcomes independently of coherence.

\subsection{Dataset}
We base our study on the WMT 2025 English$\rightarrow$Korean test set~\citep{kocmi-etal-2025-findings} in \textit{news}, \textit{social}, and \textit{literary} domains. We exclude the \textit{speech} domain, whose single-utterance composition is unsuitable for document-level analysis. Korean is known to pose substantial challenges for document-level translation, with discourse-sensitive phenomena including pro-drop, honorifics, and cross-segment entity consistency~\citep{lee-etal-2025-testset, park-pado-2024-multi, choi-etal-2018-automatic}. After filtering and length balancing (details in \Cref{appx:data}), we obtain 3{,}070 unique translations from 9 MT systems and one human reference (\mRef), across 23 documents. Each segment is evaluated under all three conditions by two annotators, yielding 18{,}420 expert judgments, matching the scale of a full WMT annotation campaign of a single language pair.

\subsection{Manipulation checks}
\label{subsec:manipulation}
The counterfactual is only informative if \docshuff really is the incoherent condition it is meant to be. We ran two separate studies, on different comparison units, that answer different questions. Details are discussed in \Cref{appx:pref_study}.

\paragraph{Do the pooled systems differ?}
The first study compares \emph{system pairs}. For the three pairs with the highest MQM rank overlap, raters saw a source segment alongside two anonymized translations and judged which was better, or whether they were indistinguishable (486 judgments, three raters). A quality difference was perceived in 85--94\% of judgments, even for these maximally overlapping pairs. The segments pooled into \docshuff are therefore genuinely different translations, not near-identical ones.

\paragraph{Does the assembled document read as inconsistent?}
Difference between systems is not inconsistency within a document, and only the second property is what \docshuff is meant to create. The second study therefore compares \emph{documents}. We sampled 102 matched passage pairs, each showing the same six contiguous source segments twice, once as \doc and once as \docshuff. Three raters, blind to the manipulation, chose which passage reads as the work of a single consistent translator. \doc was chosen in $87.3\%$ of trials ($95\%$ CI $[79.4, 92.4]$; Krippendorff's $\alpha = 0.73$; binomial $p < 0.001$). Readers asked about consistency therefore separate the two conditions reliably.

\vspace{1em}
Two consequences follow for how the results below should be read. A weak manipulation is ruled out as an explanation of any \doc--\docshuff equivalence, though other explanations are not. And a null becomes interpretable: had annotators been unable to perceive the disruption, this control would have returned chance performance. It did not. The disruption is available to be perceived, so the question is whether ESA records it.

\begin{figure*}[ht]
\centering
\includegraphics[width=\linewidth]{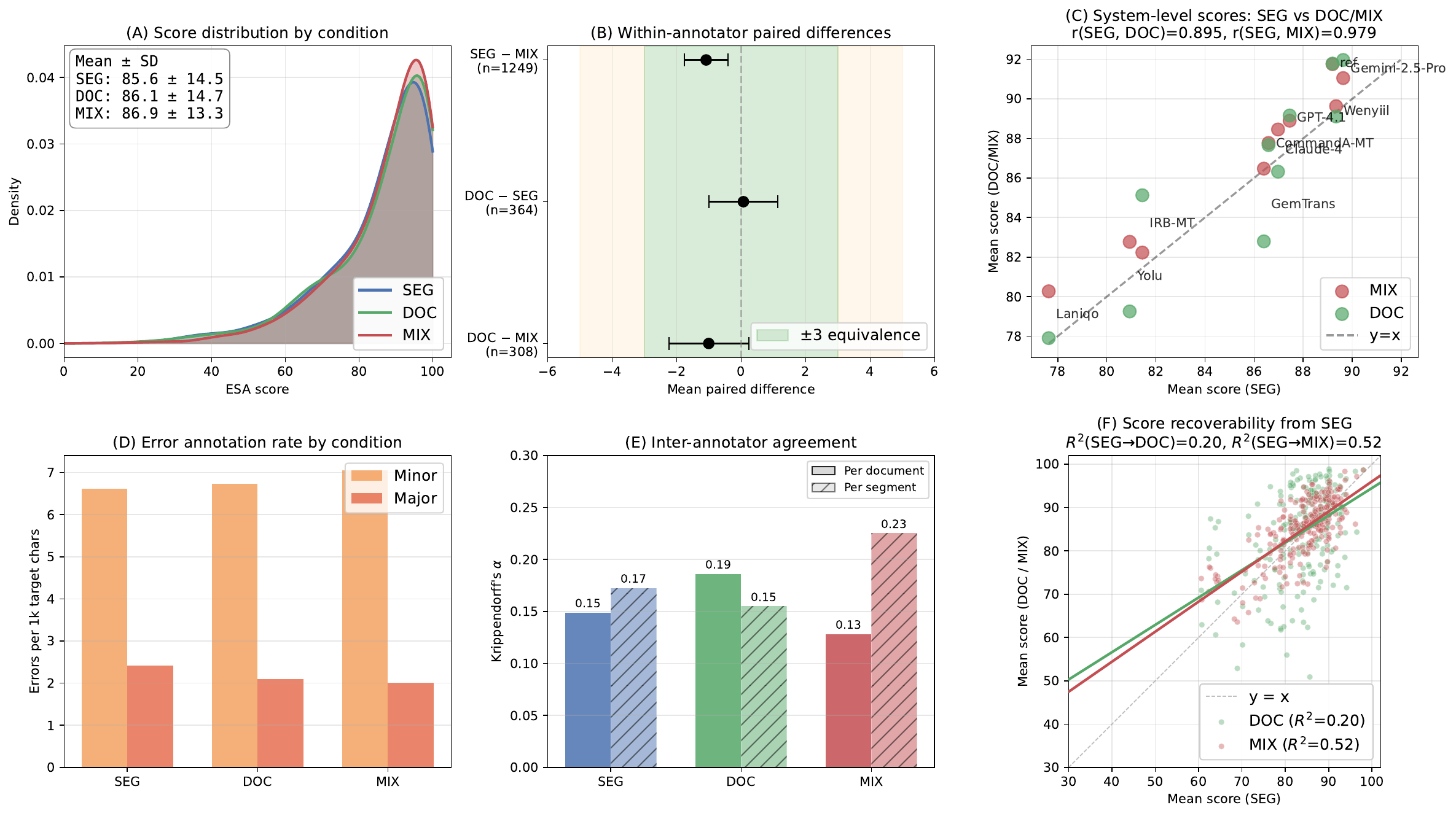}
\caption{Overview of human ESA evaluation results across three conditions (\seg, \doc, \docshuff). \textbf{(A)} Score distributions (KDE) are nearly identical across conditions; mean and SD shown in inset. \textbf{(B)} Within-annotator paired differences fall within the $\pm 3$ equivalence band for all three comparisons (\Cref{h1}); each is signed as labelled, e.g.\ \doc\ $-$ \docshuff. \textbf{(C)} System-level mean scores: \seg\ vs.\ document-level conditions; points near $y = x$ indicate that doc-level rankings closely follow segment-level rankings (\Cref{h2}). \textbf{(D)} Error annotation rate per 1{,}000 target characters by severity (\Cref{h3}). \textbf{(E)} Krippendorff's $\alpha$ computed per document and per segment. The two units yield opposite orderings: \doc\ agrees most per document but least per segment. \textbf{(F)} Score recoverability: \seg\ scores predict $52\%$ of the variance in \docshuff\ scores but only $20\%$ of \doc\ scores ($n=230$ matched pairs per regression).}
\label{fig:main_result}
\end{figure*}

\subsection{Human Evaluation}

We use the ESA annotation protocol~\citep{kocmi-etal-2024-error} on the Pearmut annotation platform~\citep{zouhar2026pearmut}. Each rater annotates 706 tasks drawn randomly from the three conditions, blinded to condition assignment. We recruited 10 professional En$\rightarrow$Ko translators, all native speakers of Korean. Further details are reported in \Cref{sec:human_eval_detail}.

\subsection{Automatic Metrics}
\paragraph{Segment-level.} \chrf~\citep{popovic-2015-chrf}; the learned metrics \xcomet~\citep{guerreiro-etal-2024-xcomet} and \metricx~\citep{juraska-etal-2024-metricx}, together with their Quality Estimation (QE)  variants \xcometkiwi~\citep{rei-etal-2023-scaling} and \metricxqe; and the LLM-as-judge metric \gemba~\citep{junczys-dowmunt-2025-gemba}.

\paragraph{Document-level.} As most discourse-targeted metrics do not support Korean, we use \dbleu~\citep{liu-etal-2020-multilingual-denoising}, \doccomet~\citep{vernikos-etal-2022-embarrassingly}, \slide~\citep{raunak-etal-2023-evaluating}, and the LLM-as-judge metric \falcon~\citep{Kim_2025-falcon}. We also include document-level variants of \xcomet and \metricx (and their QE counterparts), denoted with a \texttt{-doc} suffix.

\subsection{Testing Procedure}
\paragraph{Hypothesis testing.}
We test each hypothesis with a single targeted procedure. For \Cref{h1}, we apply the Two One-Sided Tests (TOST) procedure~\citep{schuirmann1987, Lakens2017} to assess whether \doc and \docshuff scores are statistically equivalent within $\epsilon = 3$. For \Cref{h2}, we compute Kendall's $\tau$~\citep{kendall_tau} between $r_\mathrm{DOC}$ and $r_\mathrm{MIX}$ on system-level mean scores with bootstrap confidence intervals, and compare the lower bound against $\tau_{\min} = 0.667$, the conventional two-thirds threshold for acceptable ranking agreement. Because raw rank correlations can overstate differences between statistically indistinguishable systems, we additionally cluster systems via pairwise bootstrap tests following WMT methodology~\citep{kocmi-etal-2025-findings}. For \Cref{h3}, we apply paired TOST on per-segment error counts with $\epsilon_{\mathrm{err}} = 0.5$ errors, roughly $20\%$ of the baseline rate of 2.3 errors per annotation. Full methodological details appear in \Cref{appx:evaluation}.

\paragraph{Behavioral measures.} Beyond hypothesis testing, we measure IAA (Krippendorff's $\alpha$) and per-segment annotation time as complementary evidence on annotator's cognitive efforts across conditions.

\section{Results}
\label{sec:analysis1}
\Cref{fig:main_result} summarizes our human evaluation results. The kernel density estimates (KDE) in panel (A) show that score distributions are nearly identical across all three conditions. This is the central pattern of the section: \doc and \docshuff behave equivalently under ESA evaluation, both for human annotators and for automatic metrics. The following subsections present each result by hypothesis (\Cref{sec:analysis_p1,sec:analysis_p2,sec:analysis_p3}), with the mechanism analyzed in \Cref{sec:mechanism}. Behavioral analyses and domain-wise breakdowns appear in \Cref{subsec:behavioral} and \Cref{subsec:domainwise}. \Cref{h1} and \Cref{h2} apply to both human ESA scores and automatic metric scores, and \Cref{h3} applies only to human annotations, since most metrics tested do not produce error annotations.

\subsection{[\Cref{h1}] Scores are insensitive to discourse}
\label{sec:analysis_p1}

\paragraph{Statistically equivalent scores.}
\Cref{tab:h1_combined} reports paired TOST results across human ESA scores and 14 automatic metrics. For human ESA, mean scores are nearly identical across conditions (\seg: $85.56$, \doc: $86.11$, \docshuff: $86.93$), and within-pair differences are all within $1.1$ points (\Cref{fig:main_result}B). TOST rejects the null at $\epsilon = 3$ for every pairwise comparison ($p < 0.001$), establishing equivalence within $\pm 3$ points. We note that \docshuff scores marginally exceed those of \doc, the opposite of the predicted direction, although the two conditions remain statistically indistinguishable. The same equivalence holds across all automatic metrics: mean differences between \doc and \docshuff are small relative to each metric's scale, and TOST rejects the null at the per-scale equivalence margin for every metric ($p < 0.001$).

\paragraph{Document-level metrics react in the wrong direction.}
All 14 metrics support equivalence between \doc and \docshuff (\Cref{tab:h1_combined}). Among them, the metrics with the largest raw differences are precisely the document-level ones: \metricxqe-\texttt{doc} ($d = -0.69$), \metricx-\texttt{doc} ($d = -0.63$), and \dbleu ($d = -0.28$). This negative sign which is also observed in human scores, however, means that \docshuff receives marginally higher scores than \doc, the opposite of the direction of genuine coherence sensitivity (\Cref{h1}). This reversal is more pronounced in document-level metrics than in segment-level ones. The present data do not let us identify why the reversal arises, but they do establish what ESA's outputs were not tracking. Neither the human scores nor any metric score moved with cross-segment coherence.

\subsection{[\Cref{h2}] System rankings collapse}
\label{sec:analysis_p2}

\paragraph{Rankings converge across conditions.}
\doc-based rankings are nearly indistinguishable from \docshuff-based rankings (\Cref{tab:h2_combined}). Human ESA yields $\tau = 0.778$, and every automatic metric yields $\tau = 1.00$.\footnote{The bootstrap CI for human ESA is $[0.539, 0.911]$, whose lower bound does not clear $\tau_{\min}$. See \Cref{appx:ranking}, where the clustering result is the more informative comparison.} The same holds for \seg: $\tau$ is $0.895$ between \seg and \doc, and $0.979$ between \seg and \docshuff (\Cref{fig:main_result}C). Neither document-level condition produces a system ordering distinct from segment-level evaluation.

\begin{table}[]
\centering
\resizebox{\linewidth}{!}{
\begin{tabular}{lrrrr}
    \toprule
    \textbf{Evaluator / Metric} & $n$ & $\varepsilon$ & \textbf{Mean diff} & \textbf{90\% CI} \\
    \midrule
    \multicolumn{5}{l}{\textit{Human ESA: pairwise paired TOST}} \\
    \rowcolor{yellow!15} \quad \doc\ vs \docshuff   & 308  & 3.00 & \textcolor{red!90!black}{$-1.00$} & $[-2.05, +0.04]$ \\
    \quad \doc\ vs \seg       & 364  & 3.00 & $+0.07$ & $[-0.83, +0.97]$ \\
    \quad \seg\ vs \docshuff  & 1249 & 3.00 & $-1.09$ & $[-1.65, -0.52]$ \\
    \midrule
    \multicolumn{5}{l}{\textit{Segment-level metrics: \doc\ vs \docshuff}} \\
    \quad \chrf        & 230 & 3.00 & $+0.0187$ & $[-0.515, +0.552]$ \\
    \quad \metricx    & 230 & 0.50 & $+0.0037$ & $[-0.113, +0.121]$ \\
    \quad \metricxqe & 230 & 0.50 & \textcolor{red!90!black}{$-0.0009$} & $[-0.105, +0.103]$ \\
    \quad \xcomet      & 230 & 0.05 & $+0.0002$ & $[-0.016, +0.016]$ \\
    \quad \xcometkiwi & 230 & 0.05 & $+0.0001$ & $[-0.006, +0.007]$ \\
    \quad \gemba      & 230 & 3.00 & \textcolor{red!90!black}{$-0.1538$} & $[-0.892, +0.584]$ \\
    \midrule
    \multicolumn{5}{l}{\textit{Document-level metrics: \doc\ vs \docshuff}} \\
    \quad \dbleu      & 230 & 3.00 & \textcolor{red!90!black}{$-0.2795$} & $[-0.837, +0.278]$ \\
    \quad \doccomet         & 230 & 0.05 & \textcolor{red!90!black}{$-0.0008$} & $[-0.003, +0.001]$ \\
    \quad \slide             & 230 & 0.05 & $+0.0016$ & $[-0.007, +0.010]$ \\
    \quad \metricx-\texttt{doc}    & 230 & 3.00 & \textcolor{red!90!black}{$-0.6346$} & $[-1.287, +0.018]$ \\
    \quad \metricxqe-\texttt{doc} & 230 & 3.00 & \textcolor{red!90!black}{$-0.6914$} & $[-1.350, -0.033]$ \\
    \rowcolor{gray!15} \quad \xcomet-\texttt{doc}     & 230 & 0.05 & \textcolor{red!90!black}{$-0.0119$} & $[-0.019, -0.004]$ \\
    \quad \xcometkiwi-\texttt{doc} & 230 & 3.00 & $+0.0104$ & $[-0.002, +0.023]$ \\
    \quad \falcon     & 230 & 3.00 & $+0.0421$ & $[-0.687, +0.771]$ \\

\bottomrule
\end{tabular}
}
\caption{\textbf{Paired TOST for \Cref{h1}: score equivalence across evaluators.} Human ESA reports all three pairwise comparisons; metrics report only \doc vs.\ \docshuff. TOST rejects the null at $p < 0.001$ for all rows, establishing equivalence within $\pm \varepsilon$ ($\varepsilon \in \{0.05, 0.5, 3.0\}$ depending on scale). The shaded row indicates a known-failure case for context truncation. The metrics with wrong direction are colored in red.}
\label{tab:h1_combined}
\end{table} 
\definecolor{rank1}{HTML}{A5D6A7}
\definecolor{rank2}{HTML}{C8E6C9}
\definecolor{rank3}{HTML}{DCEDC8}
\definecolor{rank4}{HTML}{F0F4C3}
\definecolor{rank5}{HTML}{FFF9C4}
\definecolor{rank6}{HTML}{FFE0B2}
\definecolor{rank7}{HTML}{FFCCBC}

\begin{table*}[ht]
\centering
\setlength{\tabcolsep}{4pt}
\resizebox{\linewidth}{!}{
\begin{tabular}{lccc|ccccccccccccc}
\toprule
& \multicolumn{3}{c|}{\textbf{Human}} 
& \multicolumn{2}{c}{\textbf{\chrf}} 
& \multicolumn{2}{c}{\textbf{\xcomet}} 
& \multicolumn{2}{c}{\textbf{\xcometkiwi}} 
& \multicolumn{2}{c}{\textbf{\metricx$^*$}} 
& \multicolumn{2}{c}{\textbf{\metricxqe$^*$}}
& \multicolumn{2}{c}{\textbf{\gemba}} \\
\cmidrule(lr){2-4} \cmidrule(lr){5-6} \cmidrule(lr){7-8} \cmidrule(lr){9-10} \cmidrule(lr){11-12} \cmidrule(lr){13-14} \cmidrule(lr){15-16}
\textbf{System} 
& {\tiny \seg} & {\tiny \doc} & {\tiny \docshuff}
& {\tiny \doc} & {\tiny \docshuff}
& {\tiny \doc} & {\tiny \docshuff}
& {\tiny \doc} & {\tiny \docshuff}
& {\tiny \doc} & {\tiny \docshuff}
& {\tiny \doc} & {\tiny \docshuff}
& {\tiny \doc} & {\tiny \docshuff}\\
\midrule
\mGemini
  & \cellcolor{rank1} 89.63 (1) & \cellcolor{rank1} 91.97 (1) & \cellcolor{rank2} 91.05 (2)
  & \cellcolor{rank1} 33.18 (1) & \cellcolor{rank1} 33.21 (1)
  & \cellcolor{rank1} .791 (1) & \cellcolor{rank1} .792 (1)
  & \cellcolor{rank3} .845 (3) & \cellcolor{rank2} .845 (\textbf{2})
  & \cellcolor{rank1} 3.83 (1) & \cellcolor{rank1} 3.81 (1)
  & \cellcolor{rank1} 3.72 (1) & \cellcolor{rank1} 3.72 (1)
  & \cellcolor{rank1} 87.95 (1) & \cellcolor{rank1} 87.91 (1) \\
\mRef
  & \cellcolor{rank1} 89.20 (1) & \cellcolor{rank1} 91.78 (1) & \cellcolor{rank1} 91.75 (1)
  & --- & --- & --- & --- & --- & --- & --- & --- & --- & --- & --- & --- \\
\textsc{Wenyiil}
  & \cellcolor{rank1} 89.34 (1) & \cellcolor{rank2} 89.11 (2) & \cellcolor{rank2} 89.63 (2)
  & \cellcolor{rank3} 30.72 (3) & \cellcolor{rank3} 30.79 (3)
  & \cellcolor{rank1} .790 (1) & \cellcolor{rank1} .787 (1)
  & \cellcolor{rank2} .852 (2) & \cellcolor{rank2} .851 (2)
  & \cellcolor{rank1} 3.97 (1) & \cellcolor{rank1} 3.97 (1)
  & \cellcolor{rank1} 3.64 (1) & \cellcolor{rank1} 3.66 (1)
  & \cellcolor{rank2} 86.74 (2) & \cellcolor{rank2} 86.71 (2) \\
\mGpt
  & \cellcolor{rank2} 87.45 (2) & \cellcolor{rank2} 89.16 (2) & \cellcolor{rank2} 88.90 (2)
  & \cellcolor{rank3} 30.35 (3) & \cellcolor{rank3} 30.25 (3)
  & \cellcolor{rank1} .782 (1) & \cellcolor{rank1} .782 (1)
  & \cellcolor{rank3} .837 (3) & \cellcolor{rank3} .836 (3)
  & \cellcolor{rank2} 4.30 (2) & \cellcolor{rank2} 4.27 (2)
  & \cellcolor{rank3} 3.98 (3) & \cellcolor{rank3} 3.97 (3)
  & \cellcolor{rank2} 86.18 (2) & \cellcolor{rank2} 86.22 (2) \\
\mClaude
  & \cellcolor{rank2} 86.98 (2) & \cellcolor{rank3} 86.32 (3) & \cellcolor{rank2} 88.45 (\textbf{2})
  & \cellcolor{rank2} 32.21 (2) & \cellcolor{rank2} 32.29 (2)
  & \cellcolor{rank3} .735 (3) & \cellcolor{rank3} .734 (3)
  & \cellcolor{rank3} .842 (3) & \cellcolor{rank3} .842 (3)
  & \cellcolor{rank2} 4.39 (2) & \cellcolor{rank2} 4.37 (2)
  & \cellcolor{rank3} 4.09 (3) & \cellcolor{rank3} 4.10 (3)
  & \cellcolor{rank3} 85.43 (3) & \cellcolor{rank3} 85.39 (3) \\
\mCommandA
  & \cellcolor{rank2} 86.59 (2) & \cellcolor{rank3} 87.66 (3) & \cellcolor{rank3} 87.77 (3)
  & \cellcolor{rank4} 29.14 (4) & \cellcolor{rank4} 29.11 (4)
  & \cellcolor{rank2} .770 (2) & \cellcolor{rank2} .772 (2)
  & \cellcolor{rank1} .861 (1) & \cellcolor{rank1} .862 (1)
  & \cellcolor{rank2} 4.26 (2) & \cellcolor{rank2} 4.27 (2)
  & \cellcolor{rank2} 3.81 (2) & \cellcolor{rank2} 3.80 (2)
  & \cellcolor{rank3} 85.07 (3) & \cellcolor{rank3} 85.12 (3) \\
\mGemtrans
  & \cellcolor{rank2} 86.40 (2) & \cellcolor{rank5} 82.80 (5) & \cellcolor{rank3} 86.47 (\textbf{3})
  & \cellcolor{rank4} 28.67 (4) & \cellcolor{rank4} 28.65 (4)
  & \cellcolor{rank1} .790 (1) & \cellcolor{rank1} .791 (1)
  & \cellcolor{rank4} .830 (4) & \cellcolor{rank4} .831 (4)
  & \cellcolor{rank1} 3.86 (1) & \cellcolor{rank1} 3.87 (1)
  & \cellcolor{rank1} 3.60 (1) & \cellcolor{rank1} 3.60 (1)
  & \cellcolor{rank4} 83.21 (4) & \cellcolor{rank4} 83.26 (4) \\
\mIrb
  & \cellcolor{rank3} 81.45 (3) & \cellcolor{rank4} 85.13 (4) & \cellcolor{rank4} 82.23 (4)
  & \cellcolor{rank6} 25.95 (6) & \cellcolor{rank6} 25.97 (6)
  & \cellcolor{rank4} .716 (4) & \cellcolor{rank3} .717 (\textbf{3})
  & \cellcolor{rank5} .806 (5) & \cellcolor{rank5} .806 (5)
  & \cellcolor{rank4} 4.83 (4) & \cellcolor{rank4} 4.82 (4)
  & \cellcolor{rank4} 4.44 (4) & \cellcolor{rank5} 4.45 (\textbf{5})
  & \cellcolor{rank5} 81.34 (5) & \cellcolor{rank5} 81.29 (5) \\
\mYolu
  & \cellcolor{rank3} 80.93 (3) & \cellcolor{rank6} 79.25 (6) & \cellcolor{rank4} 82.77 (\textbf{4})
  & \cellcolor{rank5} 27.64 (5) & \cellcolor{rank5} 27.58 (5)
  & \cellcolor{rank3} .739 (3) & \cellcolor{rank3} .738 (3)
  & \cellcolor{rank2} .848 (2) & \cellcolor{rank2} .848 (2)
  & \cellcolor{rank3} 4.50 (3) & \cellcolor{rank3} 4.50 (3)
  & \cellcolor{rank2} 3.88 (2) & \cellcolor{rank2} 3.88 (2)
  & \cellcolor{rank4} 82.86 (4) & \cellcolor{rank4} 82.90 (4) \\
\mLanqio
  & \cellcolor{rank4} 77.63 (4) & \cellcolor{rank6} 77.90 (6) & \cellcolor{rank5} 80.27 (\textbf{5})
  & \cellcolor{rank7} 23.36 (7) & \cellcolor{rank7} 23.41 (7)
  & \cellcolor{rank3} .719 (3) & \cellcolor{rank3} .721 (3)
  & \cellcolor{rank4} .833 (4) & \cellcolor{rank4} .833 (4)
  & \cellcolor{rank4} 4.86 (4) & \cellcolor{rank4} 4.84 (4)
  & \cellcolor{rank3} 4.12 (3) & \cellcolor{rank4} 4.12 (\textbf{4})
  & \cellcolor{rank5} 79.52 (5) & \cellcolor{rank5} 79.55 (5) \\
\midrule
\# clusters
  & \multicolumn{3}{c|}{4 / 6 / 5}
  & \multicolumn{2}{c}{7 / 7}
  & \multicolumn{2}{c}{4 / 3}
  & \multicolumn{2}{c}{5 / 5}
  & \multicolumn{2}{c}{4 / 4}
  & \multicolumn{2}{c}{4 / 5}
  & \multicolumn{2}{c}{5 / 5} \\
$\tau$ (\doc, \docshuff)
  & \multicolumn{3}{c|}{$0.78$}
  & \multicolumn{2}{c}{$1.00$}
  & \multicolumn{2}{c}{$1.00$}
  & \multicolumn{2}{c}{$1.00$}
  & \multicolumn{2}{c}{$1.00$}
  & \multicolumn{2}{c}{$1.00$}
  & \multicolumn{2}{c}{$1.00$} \\
\bottomrule
\end{tabular}
}
\caption{\textbf{System rankings under each evaluator for \Cref{h2}.} Cell colors indicate cluster rank (darker green = higher), derived from pairwise bootstrap tests ($B=1{,}000$, $\alpha=0.05$); systems within a cluster are not significantly distinguishable. Bold values mark cluster shifts between \doc and \docshuff. Bottom rows report the number of clusters per condition and Kendall's $\tau$ between \doc and \docshuff system rankings. $^*$(lower = better).}
\label{tab:h2_combined}
\end{table*} 

\paragraph{Statistical clusters coincide.}
For human ESA, \seg produces 4 clusters, while \doc and \docshuff produce 6 and 5. The cluster counts differ, but the compositions under \doc and \docshuff are nearly the same. The top cluster (\mGemini, \mRef) and the bottom-ranked system (\mLanqio) are shared, and 8 of 10 systems sit at identical or adjacent ranks across the two conditions. Automatic metrics show the same pattern (\Cref{tab:h2_combined}). Cluster counts vary minimally between \doc and \docshuff across all metrics. The extra granularity that \doc and \docshuff provide over \seg is therefore not driven by cross-segment coherence. Document-level presentation produces finer system distinctions whether the document is coherent or not.

\begin{table}[t]
\centering
\small
\resizebox{\linewidth}{!}{
    \begin{tabular}{lrrr}
    \toprule
    \multicolumn{4}{l}{\textit{(a) Per-annotation error counts}} \\
    \midrule
    Condition & $n$ & Total & Major \\
    \midrule
    \seg      & 5{,}920 & 2.36 & 0.63 \\
    \doc      & 5{,}920 & 2.31 & 0.55 \\
    \rowcolor{yellow!12}
    \docshuff & 5{,}920 & 2.37 & 0.53 \\
    \midrule
    \multicolumn{4}{l}{\textit{(b) Paired TOST: total errors ($\varepsilon_{\mathrm{err}} = 0.5$)}} \\
    \midrule
    Comparison & $n$ pairs & Mean diff (90\% CI) & TOST $p$ \\
    \midrule
    \rowcolor{yellow!12}
    \doc\ vs \docshuff   & 1{,}157 & \textcolor{blue}{$-0.15$} $[-0.25, -0.05]$ & $<$0.001 \\
    \doc\ vs \seg       & 1{,}184 & $+0.02$ $[-0.06, +0.11]$ & $<$0.001 \\
    \seg\ vs \docshuff  & 1{,}185 & $-0.08$ $[-0.18, +0.01]$ & $<$0.001 \\
    \bottomrule
    \end{tabular}
}
\caption{\textbf{Error counts and paired TOST results for \Cref{h3}.} (a) Mean error counts per annotation across conditions. (b) Pairwise paired TOST on total errors with $\varepsilon_{\mathrm{err}} = 0.5$. The highlighted row marks \Cref{h3}'s primary comparison. TOST rejects the null at $p < 0.001$ for all comparisons, establishing equivalence within $\pm 0.5$ errors per annotation.}
\label{tab:h3_combined}
\end{table} 

\subsection{[\Cref{h3}] No additional errors are marked}
\label{sec:analysis_p3}

\paragraph{Error counts are equivalent.}
\Cref{tab:h3_combined} reports paired TOST results for error counts. Mean error counts per annotation are nearly identical across conditions (\seg: $2.36$, \doc: $2.31$, \docshuff: $2.37$), with within-pair differences of at most $0.15$. TOST rejects the null at $\epsilon_{\mathrm{err}} = 0.5$ errors per annotation ($p < 0.001$) for every comparison, showing equivalence within roughly $20\%$ of the baseline error rate. For Major errors, the direction even reverses: \docshuff shows fewer Major errors than \doc ($0.53$ vs.\ $0.55$; \Cref{fig:main_result}D), opposite to the predicted direction.

\paragraph{Error-free rates coincide.}
The proportion of error-free annotations is nearly identical between \doc ($15.5\%$) and \docshuff ($15.7\%$), with \seg slightly lower at $12.8\%$. Documents in \docshuff are marked as error-free as often as coherent ones, despite their deliberate incoherence.

\paragraph{Error spans are not boundary-localized.}
Signal \circled{3} predicts that spans should concentrate where consecutive segments meet, since that is where \docshuff's discontinuities occur. They do not. Counting a span as boundary-localized when it falls within the first or last $20\%$ of its segment, the rate is $50.3\%$ under \seg, $50.1\%$ under \doc and $50.0\%$ under \docshuff ($n = 49{,}749$ spans; $\docshuff - \doc = -0.1$ points, $95\%$ CI $[-1.2, +1.0]$). The pattern is unchanged when zero-width omission markers are excluded, when the analysis is restricted to document-interior segments, and when it is restricted to \textit{Major} spans (\Cref{appx:errors}). Signal \circled{3} therefore fails alongside \Cref{h1,h2,h3}: whatever annotators noticed at the seams, they did not mark it where the seams are.

\subsection{Mechanism: \docshuff is evaluated in isolation}
\label{sec:mechanism}
The preceding hypotheses show that ESA fails to distinguish \doc from \docshuff. Two analyses indicate why: under \docshuff, both human annotators and automatic metrics fall back to segment-level evaluation. We establish this through score recoverability for human annotators and through ranking agreement for automatic metrics.

\paragraph{Score recoverability.}
We fit linear regressions predicting each document-level score from the corresponding \seg score, matched by (\texttt{doc\_id}, \texttt{system}) ($n = 230$ matched pairs per regression; \Cref{fig:main_result}F). \seg scores explain $52\%$ of the variance in \docshuff scores ($R^2 = 0.52$, $r = 0.72$), but only $20\%$ of \doc scores ($R^2 = 0.20$, $r = 0.45$). The higher $R^2$ for \docshuff indicates that, even with the full document on screen, annotators fall back to segment-level judgments once inter-sentence coherence is broken. When document-level cues become uninformative, each segment is rated much as it would be in isolation.

\begin{figure} 
    \centering
    \includegraphics[width=1\linewidth]{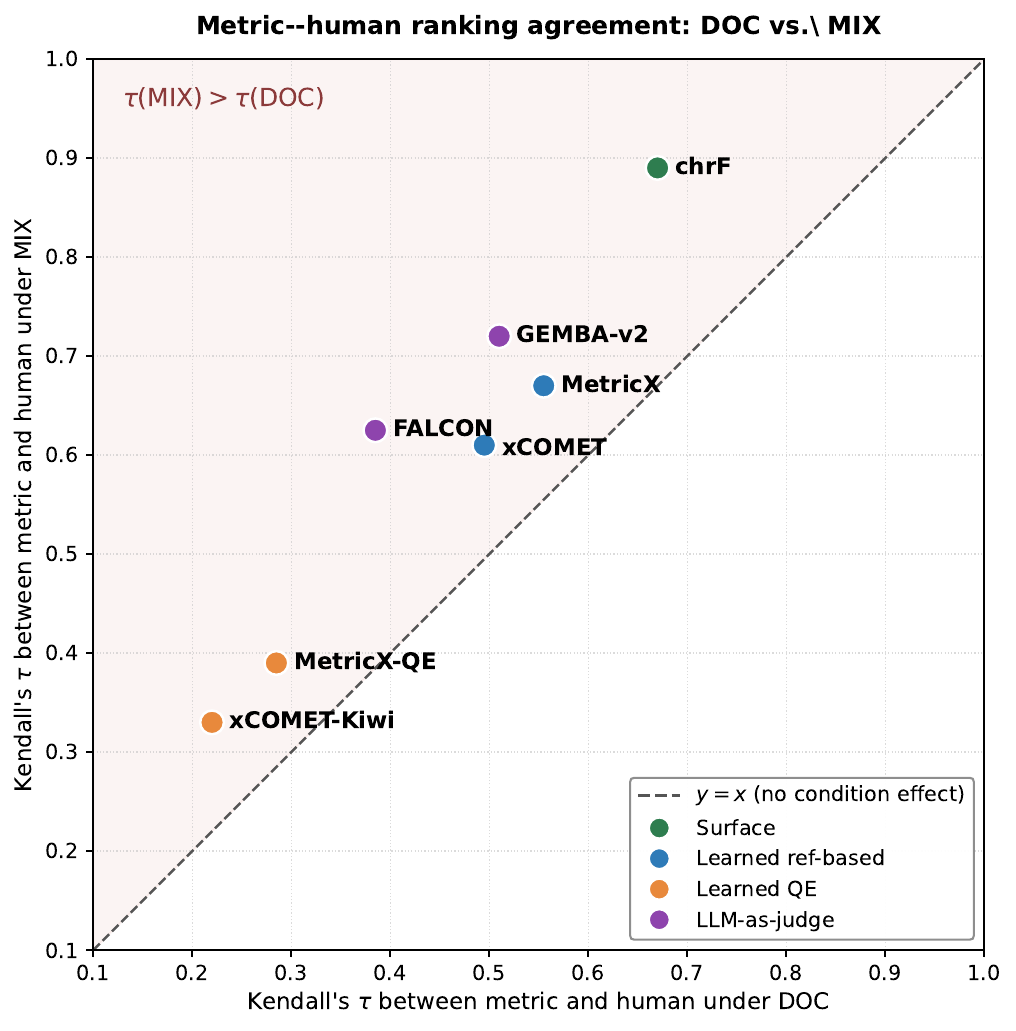}
    \caption{\textbf{Metric$\bm{\leftrightarrow}$human ranking agreement under \doc and \docshuff.} Each point shows Kendall's $\tau$ between a metric's system ranking and the human ranking, under \doc ($x$-axis) and \docshuff ($y$-axis). All metrics lie above $y = x$ (shaded region), meaning that every metric agrees more closely with human rankings under \docshuff than under \doc.
    }
    \label{fig:metric_corr}
\end{figure}

\paragraph{Metric-human agreement peaks under \docshuff.}
\Cref{fig:metric_corr} shows the same pattern in human-metric system correlations: every metric of every category, from surface to learned to LLM-as-judge, lies above $y = x$. The peak occurs under the condition in which humans themselves fall back to segment-level judgments, suggesting that current metrics align with humans through a shared segment-level fallback rather than through shared sensitivity to document-level quality. The effect is largest for \chrf, the most surface-level metric in the set, and smaller for learned QE metrics, which deviate least from their \doc behavior. Per-metric details appear in \Cref{appx:metric_result}.

\subsection{Behavioral analysis}
\label{subsec:behavioral}

\paragraph{IAA.}
\label{sec:analysis_p4}
\Cref{fig:main_result}E reports Krippendorff's $\alpha$ at two units of analysis, with an inverted pattern between them. Computed per document, \doc leads ($\alpha = 0.19$, vs.\ $0.15$ for \seg and $0.13$ for \docshuff). Computed per segment, the order reverses, with \docshuff leading ($0.23$ vs.\ $0.17$ for \seg and $0.15$ for \doc). The per-document result for \docshuff is constrained by construction, since \docshuff documents differ across annotators by design (\Cref{appx:iaa}).

This bears on an alternative reading of \Cref{sec:mechanism}, on which the extra variance in \doc scores is annotator noise rather than context use. Noise would depress agreement at \emph{both} units. Instead \doc agrees most per document and least per segment: annotators diverge on individual segments yet converge on overall impressions, which is the signature of context processed and expressed idiosyncratically rather than of random noise. IAA does not directly measure context use, so we treat this as corroborating rather than decisive.

\begin{figure*} 
    \centering
    \includegraphics[width=1\linewidth]{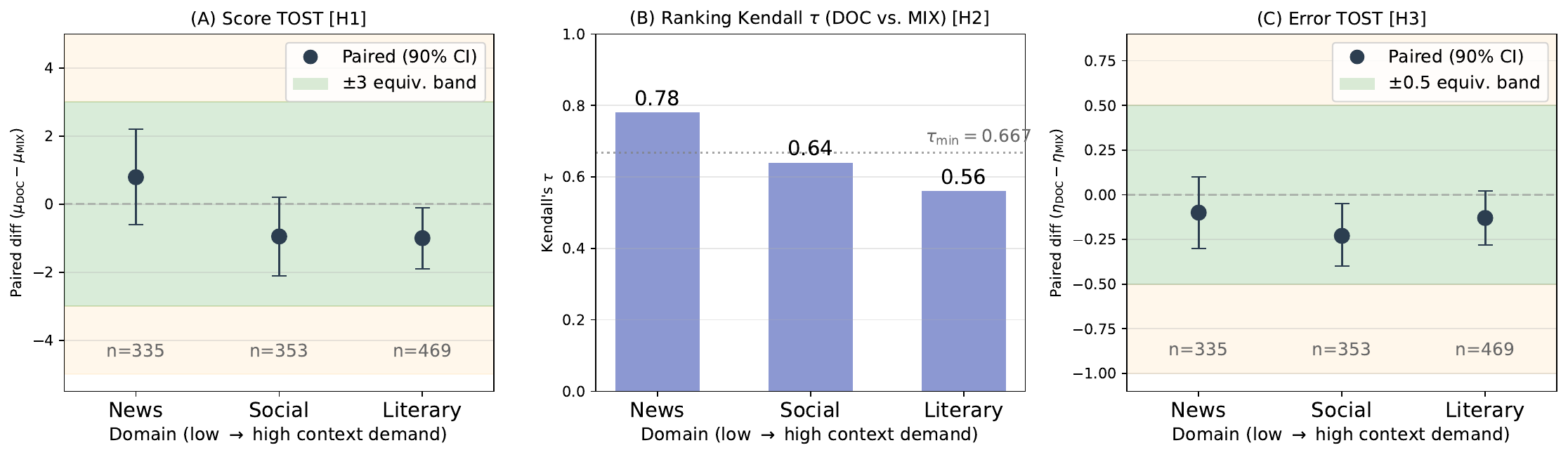}
    \caption{Domain-stratified replication of \Cref{h1,h2,h3} on human ESA, with domains ordered by increasing context demand (news $<$ social $<$ literary). Throughout, the plotted paired difference is \doc\ $-$ \docshuff. \textbf{(A)} Score TOST: paired mean differences on segment scores. All point estimates fall within the $\pm 3$ equivalence band. \textbf{(B)} System ranking agreement between \doc\ and \docshuff. The dotted line marks the conventional acceptability threshold ($\tau_{\min} = 0.667$). \textbf{(C)} Error TOST: paired mean differences on per-annotation error counts. All estimates fall within the $\pm 0.5$ equivalence band.}
    \label{fig:domainwise}
\end{figure*}

\paragraph{Annotation time.}
Annotators spend less time per segment under document-level conditions than under \seg ($61$s for \doc and $56$s for \docshuff, against $76$s for \seg; \Cref{fig:time_analysis}). The decrease suggests that document context accelerates processing whether or not the context is coherent. This fits the segment-level fallback observed in \Cref{sec:mechanism}. When annotators evaluate at the segment level, surrounding context, coherent or not, acts as a processing shortcut rather than as new information to evaluate.

\begin{table}[t]
\centering
\small
\begin{tabular}{l>{\centering}p{1cm}>{\centering}p{1cm}>{\centering\arraybackslash}p{1cm}}
\toprule
{Domain} & \seg\ & \doc\ & \hspace{-3mm}\docshuff\ \hspace*{-2mm} \\
\midrule
Literary & \cellcolor{lightgray!67} 67s & \cellcolor{lightgray!53} 53s & \cellcolor{lightgray!41} 41s \\
News & \cellcolor{lightgray!92} 92s & \cellcolor{lightgray!78} 78s & \cellcolor{lightgray!76} 76s \\
Social & \cellcolor{lightgray!74} 74s & \cellcolor{lightgray!57} 57s & \cellcolor{lightgray!60} 60s \\
\\[-0.5em]
Overall & \cellcolor{lightgray!76} 76s & \cellcolor{lightgray!61} 61s & \cellcolor{lightgray!56} 56s \\
\bottomrule
\end{tabular}
\caption{Per-segment annotation time across conditions, by domain. The decrease from \seg\ to \doc\ and \docshuff\ suggests that context shapes annotators' processing, regardless of whether that context is coherent or not.}
\label{fig:time_analysis}
\end{table}

\subsection{Domain-wise replication}
\label{subsec:domainwise}
We replicate the analyses across our three domains, ordered by an intuitive (and unmeasured) notion of context demand: literary highest, where narrative voice and character continuity carry across a document; news lowest, with standardized tone and self-contained documents; social in between. Under \Cref{h1,h2,h3}, larger \doc-vs-\docshuff differences should emerge where cross-sentential dependence is stronger.

\Cref{fig:domainwise} shows that the equivalence pattern holds across all three domains, with negligible effect sizes ($|d| < 0.13$ across all paired comparisons). We observe no graded effect. \Cref{h1} differences range from $+0.79$ in news (the only domain showing the predicted direction) to $-1.00$ in literary, with no monotonic relationship (Panel A). \Cref{h3} effects are largest in social rather than literary (Panel C). The one signal that aligns with the intuitive ordering is system ranking agreement (Panel B), where $\tau_{\text{\tiny DOC},\text{\tiny MIX}}$ decreases monotonically from news ($0.78$) through social ($0.64$) to literary ($0.56$, below the threshold $\tau_{\min} = 0.667$). Bootstrap intervals on these values qualify the ordering. Resampling segments with replacement ($B = 10{,}000$) gives $[0.556, 0.867]$ for news, $[0.378, 0.778]$ for social, and $[0.422, 0.689]$ for literary. The three intervals overlap substantially, and a value near $0.56$ falls inside all of them, and social likewise falls below $\tau_{\min}$ in the majority of resamples. The monotonic ordering in Panel B is therefore not resolvable at this sample size.\footnote{Intervals here are computed over segments rather than over source documents as in \Cref{appx:ranking}, since the literary domain rests on two source documents.}

We read this cautiously: literary rests on two source documents (\Cref{tab:dataset_stats}), and a coherence-sensitivity account predicts several signals moving together, whereas here only $\tau$ moves, with the score difference within margin and reversed in direction. A single moving signal on two documents is better explained by sampling instability. What the replication establishes is that the equivalence holds in all three domains, not that context demand has no effect anywhere.

\section{Conclusion}
\label{sec:conclusion}
Since the ALPAC report~\citep{alpac1966}, MT evaluation has been built, largely tacitly, around segment-level human judgments. Decades of gold data have accumulated under this paradigm, shaping the metrics subsequently trained or calibrated against it. Our counterfactual experiment shows what this legacy actually captures. Across human ESA and 14 automatic metrics, including discourse-targeted variants and LLM-based scorers, we find no evaluator that distinguishes coherent documents from deliberately incoherent ones, in any of the three domains tested.

Two explanations can be set aside. The first is perception: readers asked about consistency separate the two conditions reliably, and document presentation measurably changes how annotators work. The second is context length: the document-level metrics we tested received the full document and still failed, and the LLM-based scorers, with effectively unbounded context, fared no differently. What is missing is a place to put the judgment, both a notion of what constitutes document-level quality in the gold data that drives the field and an interface that elicits it.

The implication is direct. Training or fine-tuning translation models or metrics on segment-based gold is unlikely to produce document-level competence. It can, at best, reproduce the segment-level ceiling that this gold encodes. Genuine progress in document-level MT will likely require new evaluation protocols designed to elicit document-level judgments rather than merely to display document-level context, and new gold data collected under them. Until then, what is reported as document-level progress is, under our measure, hard to separate from segment-level progress.

\section*{Limitations}

\paragraph{Language selection.}
Our investigation prioritizes depth over breadth, with 18,420 annotations on a single language pair where document-level signals are strongest.
Korean's discourse-sensitive features (pro-drop, honorifics, cross-segment entity tracking) make it a setting where ESA, if it registers coherence
at all, should do so most readily. A null in this most-favorable setting bounds what can reasonably be expected elsewhere, but it does not establish
the same result elsewhere. Our finding is scoped to English$\rightarrow$Korean under ESA, and replication across typologically diverse language pairs
is the primary next step rather than a formality.

\paragraph{Domain composition.}
The three domains are unevenly represented: literary contributes $44.6\%$ of all annotations but rests on only 2 source documents
(\Cref{tab:dataset_stats}). Domain-stratified results (\Cref{subsec:domainwise}) should therefore be read as replication checks rather than as
a domain comparison, and the literary ranking result in particular is reported with bootstrap intervals for this reason.

\paragraph{Sample sizes and IAA.}
Our analyses are based on $10$ annotators and $9$ MT systems, yielding $\tbinom{9}{2} = 36$ pairwise comparisons for ranking statistics. Krippendorff's $\alpha$ values are modest ($\alpha < 0.25$ across all conditions), reflecting both the subjectivity of fine-grained translation quality assessment and the small annotator pool. We mitigate small-sample limitations with within-annotator paired designs, bootstrap confidence intervals, and statistical clustering (\Cref{appx:ranking}). 

\paragraph{Scope of the counterfactual.}
Our \docshuff condition breaks document coherence by combining segments from multiple MT systems, a strong but single form of incoherence. What it perturbs is consistency: register, terminology, and entity form. It leaves source content and sentence order intact, and it cannot produce errors internal to a single system's output, such as a mistranslated anaphor or a wrong pro-drop resolution. \docshuff is thus best understood as an upper bound on the consistency signal rather than as a sample of naturally occurring document-level errors. The manipulation checks (\Cref{tab:pref_study}) rule out a weak manipulation as an explanation of the equivalence, though other explanations remain open. Whether ESA would register subtler perturbations, such as random reordering or gradual degradation, is untested.

\paragraph{Scope of contribution.}
We focus on ESA, a current standard protocol in WMT-style MT evaluation. Whether the same failure extends to other protocols, such as MQM with explicit discourse-level error categories, is an open question. We diagnose the failure but do not propose a replacement; designing an interface that elicits document-level judgments is itself a research program.

\section*{Acknowledgment}
This work was supported by the G-LAMP Program of the National Research Foundation of Korea (NRF) grant funded by the Ministry of Education (No. RS-2025-25441317); the Ministry of Science and ICT (MSIT), and the National IT Industry Promotion Agency (NIPA) through the Advanced GPU Utilization Support Program (02-26-01-0499). This work was partly supported by the Institute of Information \& Communications Technology Planning \& Evaluation(IITP)-Innovative Human Resource Development for Local Intellectualization program grant funded by the Korea government(MSIT) (IITP-2026-RS-2022-00156360).

\section*{Ethics Statement}
Our study involves expert human annotators. Participation was voluntary, and annotators were compensated at a fair market rate of \$40 per hour. All annotations were anonymized, and no personally identifiable information was collected.

\paragraph{Licenses of artifacts.}
All datasets, models, and software used in this study are publicly available for research purposes. The WMT25 test set is distributed under CC-BY 4.0; the automatic metrics \xcomet, \metricx, \chrf, and \doccomet are released under Apache 2.0. The LLM-as-judge metrics \gemba and \falcon are computed via the OpenAI API (\mGpt) under OpenAI's terms of service. Pearmut is available under MIT. Our 18{,}420 ESA annotations are released under CC-BY 4.0 at \url{https://huggingface.co/datasets/trotacodigos/esa-counterfactual-enko}, and the analysis code under MIT at \url{https://github.com/trotacodigos/esa-counterfactual.git}.

\paragraph{Intended use.}
Our use of the WMT25 test set, automatic metrics, and the ESA protocol is consistent with their intended use for MT evaluation research. The annotations and datasets we release are intended for research use only, consistent with the access conditions of the source data.

\paragraph{Use of AI assistants.}
We acknowledge the use of AI assistants (Claude Opus 4.6) for writing refinement and code review during paper preparation. All experimental design, analysis, and claims are the authors' own.

\bibliography{anthology-2,custom,arxiv/arxiv_bib}

\newpage
\appendix
\crefalias{section}{appendix}
\crefalias{subsection}{appendix}
\crefalias{subsubsection}{appendix}

\section*{Appendix}

\section{Experimental Setup}

\subsection{Dataset}
\label{appx:data}

\paragraph{Source data.}
The Conference on Machine Translation (WMT) has organized shared tasks for MT research since 2006~\citep{koehn-monz-2006-manual}. The General Machine Translation shared task, WMT's central event, releases parallel test sets across multiple language pairs and domains each year, against which both academic and industrial systems are evaluated under standardized human and automatic protocols. Test documents are curated from authentic sources (news articles, social media, literature, speeches) and translated independently by all participating systems, with professional reference translations provided. The multi-system parallel format enables direct comparison of system outputs on identical source material, which our counterfactual evaluation design requires. We use the WMT 2025 edition~\citep{kocmi-etal-2025-findings}, which is the most recent and explicitly targets document-level evaluation.

\paragraph{Statistics.}
\label{appx:data_stats}

\Cref{tab:dataset_stats} summarizes the final dataset statistics by domain. The 10 systems comprise 9 MT systems (\mClaude, \mGpt, \mGemini, \mGemtrans, \mYolu, \mWenyiil, \mLanqio, \mIrb, \mCommandA) and one human reference (\mRef). The domain composition is unbalanced, reflecting WMT 2025's natural distribution. We address this through per-domain replication and length normalization, confirming that aggregate findings hold within each domain and across document lengths.

For statistical comparisons across conditions (\Cref{sec:analysis1}), we restrict analysis to segments evaluated under all three conditions, yielding $n = 5{,}920$ annotations per condition. For paired analyses, the number of paired observations varies by comparison: $n = 308$ (\doc vs.\ \docshuff), $n = 364$ (\doc vs.\ \seg), and $n = 1{,}249$ (\seg vs.\ \docshuff). The variation reflects different overlap structures of (annotator, segment) tuples across condition pairs.

\begin{table}[t]
\centering
\small
\resizebox{\linewidth}{!}{
\begin{tabular}{lrrrr}
\toprule
 & \textbf{News} & \textbf{Social} & \textbf{Literary} & \textbf{Total} \\
\midrule
\multicolumn{5}{l}{\textit{Source documents and segments}} \\
\# documents          & 13      & 8       & 2       & 23 \\
\# unique segments    & 80     & 90     & 137 & 307 \\
Mean segs/doc        & 6.2    & 11.3   & 68.5   & 13.4 \\
\midrule
\multicolumn{5}{l}{\textit{System outputs}} \\
\# MT systems        & 9       & 9       & 9       & 9 \\
\# human references  & 1       & 1       & 1       & 1 \\
\# unique translations & 800  & 900  & 1{,}370 & \textbf{3{,}070} \\
\midrule
\multicolumn{5}{l}{\textit{Annotations}} \\
\# evaluations & 2{,}400 & 2{,}700 & 4{,}110 & 9{,}210 \\
\# total annotations                        & 4{,}800 & 5{,}400 & 8{,}220 & \textbf{18{,}420} \\
\bottomrule
\end{tabular}
}
\caption{Dataset statistics by domain. \textit{Mean segs/doc} is \#~unique segments divided by \#~documents, counted after the sub-document splitting described in \Cref{sec:data_processing}. Each unique translation is evaluated under three conditions (\seg, \doc, \docshuff) and annotated by two independent annotators, yielding 18{,}420 total annotations. Note the unbalanced composition: the literary domain rests on 2 source documents but contributes $44.6\%$ of all annotations.}
\label{tab:dataset_stats}
\end{table}

\paragraph{Preprocessing.}
\label{sec:data_processing}

We apply the following pipeline to prepare segments for human evaluation. We first prioritize segments expected to yield the most informative human judgments based on their automatic evaluation scores, following \textsc{\small subset2eval}~\citep{subset2eval}, and select the documents that contain them as whole documents. Segments shorter than 10 tokens are appended to their immediately preceding segment, with paragraph boundaries (\verb|\n\n|) preserved throughout. The 23 documents vary substantially in length, particularly between literary documents (long-form narrative) and news or social documents (short articles or posts). To reduce length-driven variance in document-level evaluation, we split longer documents into sub-documents that approximate the cross-domain mean length. Segments exceeding the domain-average token length (80 tokens in the WMT setting) are split at the nearest line break (\verb|\n|) to preserve source--target alignment. If no line break exists within the segment, the segment is retained at its original length. Each document in the resulting corpus contains at least two segments, so that document-level context is available for evaluation.

\begin{table*}[t]
\centering
\small
\setlength{\tabcolsep}{5pt}
\begin{tabular}{llcrrrcc}
\toprule
\textbf{System Pair (AutoRank)} & \textbf{Split} & $n$ & \textbf{m1} & \textbf{m2} & \textbf{Tie} & \textbf{Sign Test} & \textbf{Distribution} \\
\midrule
\multirow{5}{*}{\shortstack[l]{\mRef{} vs. \textsc{Gemini-2.5-Pro}\\(1--3 / 1--3)}}
    & Overall (Raw)        & 162 & 35.2 & 50.0 & 14.8 & \multirow{2}{*}{$p = .050$} & \distbar{35.2}{50.0}{14.8} \\
    & Overall (Normalized) & 162 & 38.5 & 50.0 & 11.5 &                             & \distbar{38.5}{50.0}{11.5} \\
\cmidrule(l){2-8}
    & \quad Literary       & 126 & 32.5 & 50.0 & 17.5 & --- & \distbar{32.5}{50.0}{17.5} \\
    & \quad News           &  15 & 53.3 & 40.0 &  6.7 & --- & \distbar{53.3}{40.0}{6.7}  \\
    & \quad Social         &  21 & 38.1 & 57.1 &  4.8 & --- & \distbar{38.1}{57.1}{4.8}  \\
\midrule
\multirow{5}{*}{\shortstack[l]{GPT-4.1 vs. \textsc{Claude-4}\\(4--6 / 4--7)}}
    & Overall (Raw)        & 162 & 54.3 & 39.5 &  6.2 & \multirow{2}{*}{$p = .062$} & \distbar{54.3}{39.5}{6.2}  \\
    & Overall (Normalized) & 162 & 62.7 & 31.3 &  6.0 &                             & \distbar{62.7}{31.3}{6.0}  \\
\cmidrule(l){2-8}
    & \quad Literary       & 126 & 47.6 & 46.0 &  6.3 & --- & \distbar{47.6}{46.0}{6.3}  \\
    & \quad News           &  15 & 80.0 & 20.0 &  0.0 & --- & \distbar{80.0}{20.0}{0.0}  \\
    & \quad Social         &  21 & 76.2 & 14.3 &  9.5 & --- & \distbar{76.2}{14.3}{9.5}  \\
\midrule
\multirow{5}{*}{\shortstack[l]{\textsc{GemTrans} vs. \textsc{Wenyiil}\\(5--10 / 5--12)}}
    & Overall (Raw)        & 162 & 39.5 & 48.1 & 12.3 & \multirow{2}{*}{$p = .275$} & \distbar{39.5}{48.1}{12.3} \\
    & Overall (Normalized) & 162 & 40.3 & 43.8 & 15.9 &                             & \distbar{40.3}{43.8}{15.9} \\
\cmidrule(l){2-8}
    & \quad Literary       & 126 & 38.9 & 51.6 &  9.5 & --- & \distbar{38.9}{51.6}{9.5}  \\
    & \quad News           &  15 & 40.0 & 13.3 & 46.7 & --- & \distbar{40.0}{13.3}{46.7} \\
    & \quad Social         &  21 & 42.9 & 52.4 &  4.8 & --- & \distbar{42.9}{52.4}{4.8}  \\
\bottomrule
\end{tabular}
\caption{Pairwise preference results across three system pairs (486 judgments total; 162 per pair, 3 annotators). For each pair, we report Overall results (Raw and Normalized via domain-weighted averaging with a 50\% cap) and per-domain breakdowns. Distribution bars show \textcolor{blue!60}{m1 preference}, \textcolor{red!50}{m2 preference}, and \textcolor{gray!60}{tie}. Sign tests apply to non-tie judgments at the overall (raw) level only.}
\label{tab:pref_study}
\end{table*}

\subsection{Construction of \docshuff}
\label{appx:pref_study}

\paragraph{Sampling procedure.} For a source document $d$ with segments $s_1, s_2, \ldots, s_n$, the \doc condition presents all $n$ segments translated by a single system. The \docshuff condition uses the same $n$ source segments in the same order, but each segment translation is randomly drawn from the pool of system outputs for that segment. The sampling is constrained so that every system in the pool contributes at least one segment within a \docshuff document, and each segment prefers a system other than the one used at the same position under \doc. The resulting document preserves source content and sentence order while introducing system-level variation across consecutive segments, breaking discourse coherence by construction. We use a fixed random seed for reproducibility. Tasks are batched with a maximum of 300 segments each to balance annotator workload.

\paragraph{Study 1: do the pooled systems differ?}
The first study compares \emph{system pairs}. To verify that the systems pooled in \docshuff produce distinguishable translations, we conducted a pairwise preference study on the three system pairs with the highest overlap in MQM ranks (computed by AutoRank;~\citealp{kocmi-etal-2025-findings}): \mRef vs. \mGemini (both rank 1--3), \mGpt vs. \mClaude (rank 4--6 vs. 4--7), and \mGemtrans vs. \mWenyiil (rank 5--10 vs. 5--12). If raters can reliably discriminate quality for these maximally overlapping pairs, systems with larger rank separation should be at least as distinguishable.

For each pair, we sampled one document per domain (54 source segments total). Each item presented the source alongside two anonymized translations in randomized order. Three undergraduate linguistics majors independently labeled each item as ``A is better,'' ``B is better,'' or ``No difference,'' producing 486 judgments. We report both raw results and domain-normalized results (sample-size-weighted averaging with a 50\% cap), since the literary domain accounts for 78\% of segments.

We assess discriminability primarily through pooled tie rates (the proportion of ``No difference'' judgments), with per-rater tie rates as a transparency check on consistency. We additionally apply a two-sided sign test to non-tie judgments for directional preference, though this is secondary to the discriminability question.

\Cref{tab:pref_study} reports the results. Across all three pairs, raters perceived a quality difference in 85--94\% of judgments, with per-rater tie rates converging across the three raters (\mGpt vs. \mClaude: 9.3/0.0/9.3\%; \mGemtrans vs. \mWenyiil: 22.2/3.7/11.1\%; \mRef vs. \mGemini: 16.7/11.1/16.7\%) and all three ranking the pairs in the same order of discriminability. The sign test is borderline for \mRef vs. \mGemini ($p = .050$) and non-significant elsewhere, but it targets directional preference rather than discriminability: a non-significant result implies that raters did not consistently favor one system, not that they could not distinguish the two.

Since even the maximally overlapping pairs are distinguishable in at least 85\% of judgments, the systems pooled in \docshuff span a quality range broad enough that \docshuff combines genuinely different translations rather than near-identical ones. We stress what this does not establish. Difference is a property of the pooled systems, not of the assembled document, and a reader who can tell two translations apart need not experience their concatenation as inconsistent. The second study addresses that question directly.

\paragraph{Study 2: does the assembled document read as inconsistent?}
The second study compares \emph{documents}. We sampled 102 passage pairs, each showing the same six contiguous source segments twice, once as \doc and once as \docshuff. The \doc system for a pair is always one of the systems contributing to that pair's \docshuff version, so the two members are approximately matched on mean segment quality and differ mainly in whether a single system produced the whole passage. Presentation order was randomized and raters were told nothing about how the passages were constructed. Three raters (researchers in the field) chose for each pair which passage reads as the work of a single consistent translator.

\doc was chosen in $89$ of $102$ pairs ($87.3\%$, $95\%$ CI $[79.4, 92.4]$), with Krippendorff's $\alpha = 0.73$ across the three raters and a two-sided binomial test against chance at $p < 0.001$. Because the two members of a pair are approximately quality-matched by construction, this preference is unlikely to be explained by \docshuff being worse on average. It tracks within-passage consistency.

\paragraph{What \docshuff does and does not perturb.}
\Cref{tab:mix_example} illustrates the mechanism. Korean marks speech level on every sentence-final verb, so a change of contributing system mid-passage is audible to a native reader. In the passages we inspected, \docshuff also produced terminology drift and entity-form inconsistency within a passage. Errors internal to a single system's output, such as a mistranslated anaphor or a wrong pro-drop resolution, are not created by the manipulation, but neither are they removed by it: they occur in \doc and \docshuff alike, as does the source content and its sentence order. The one dimension on which the two conditions differ is within-document consistency, and \docshuff perturbs it densely and deliberately. That is what makes the null informative: a protocol blind to this is unlikely to register the sparser forms that arise naturally.

\begin{table}[t]
\centering
\small
\setlength{\tabcolsep}{3.5pt}
\renewcommand{\arraystretch}{1.25}
\begin{tabular}{@{}l l >{\raggedright\arraybackslash}p{0.50\linewidth} l@{}}
\toprule
 & \textbf{System} & \textbf{Translation} & \textbf{Level} \\
\midrule
\multicolumn{4}{@{}l}{\textit{\doc: single system throughout}} \\
$s_5$ & \textsc{gemini} & \dots 거의 같은 \textbf{높이다.} & Plain \\
$s_6$ & \textsc{gemini} & \dots 손을 흔들어 \textbf{주었다.} & Plain \\
$s_7$ & \textsc{gemini} & \dots 막지 \textbf{않는다.} & Plain \\
\addlinespace[3pt]
\midrule
\multicolumn{4}{@{}l}{\textit{\docshuff: three systems}} \\
$s_5$ & \textsc{gemini} & \dots 거의 같은 \textbf{높이다.} & Plain \\
$s_6$ & \textsc{gpt} & \dots 계속 손을 \textbf{흔들었습니다.} & \textbf{Formal.p} \\
$s_7$ & \mIrb & \dots 가리지 \textbf{않네요.} & \textbf{Informal.p} \\
\bottomrule
\end{tabular}
\caption{Three consecutive segments of one social-media document. Korean marks speech level on every sentence-final verb (bolded). \doc holds a single level throughout, whereas \docshuff moves Plain (한다체) $\rightarrow$ Formal-polite (합쇼체) $\rightarrow$ Informal-polite (해요체) within three sentences, a register discontinuity with no counterpart in the source. Both passages were shown in the same document-level ESA interface, and \docshuff was \emph{not} scored lower.}
\label{tab:mix_example}
\end{table}

\subsection{Human Annotation}
\label{sec:human_eval_detail}

\paragraph{Recruitment \& compensation.}
We worked with a recruitment vendor to identify and contract a pool of raters. After a qualification step, the vendor set up an anonymized Slack channel for ongoing communication. Prior to the main annotation, we held an onboarding session via Zoom to distribute a Korean annotation guideline and address questions. Raters were then given one week to complete a warm-up annotation, after which we provided individual feedback on the balance and consistency of their annotations relative to other raters. Raters were compensated at an industry-competitive rate of \$40 per hour.

\paragraph{Data collection.}
Main annotation was conducted from March to April 2026. To ensure sufficient within-rater coverage while limiting fatigue, raters were asked to annotate regularly over the one-month period, averaging approximately 1 to 2 hours per day. Progress was monitored daily for attention checks, with intervention when over- or under-pacing was detected. After collection, we screened annotations for outliers and requested re-annotation for flagged items where appropriate.

\begin{figure}[t]
    \centering
    \begin{subfigure}{\linewidth}
        \centering
        \includegraphics[width=\linewidth]{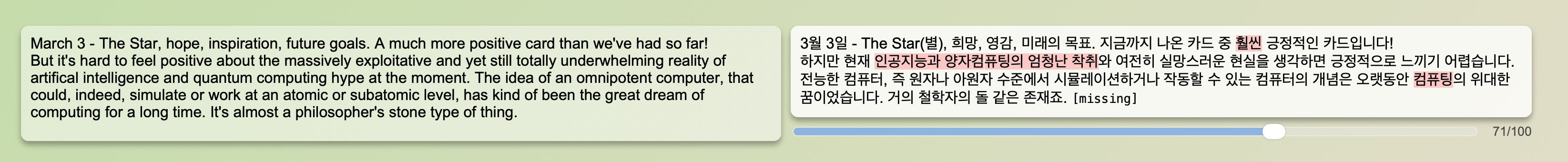}\vspace{-2mm}
        \caption{Segment-level}
        \label{fig:seg_level}
    \end{subfigure}
    \vspace{-2mm}
    
    \begin{subfigure}{\linewidth}
        \centering
        \includegraphics[width=\linewidth]{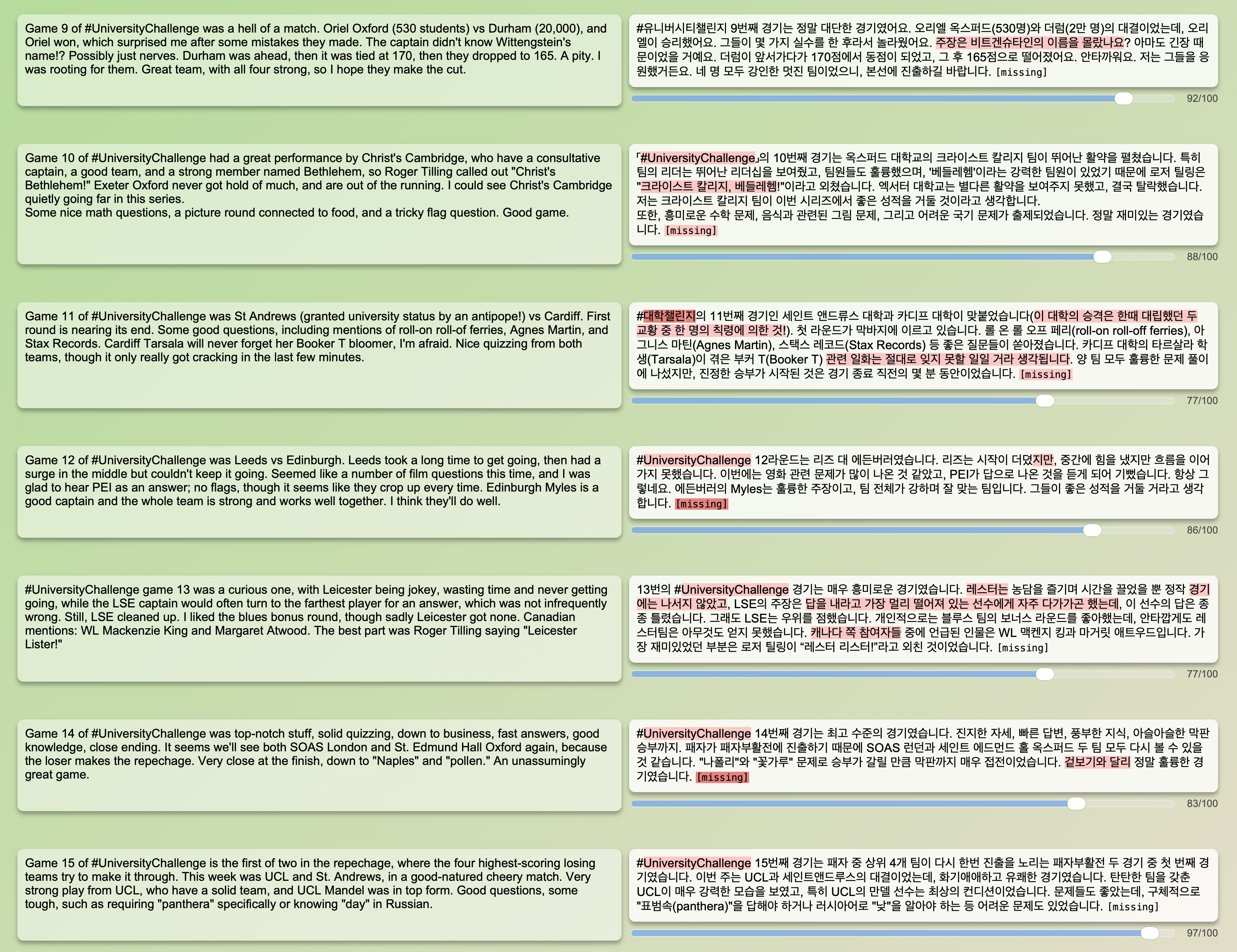}\vspace{-2mm}
        \caption{Document-level}
        \label{fig:doc_level}
    \end{subfigure}
    \caption{The segment- and document-level evaluation screens of Pearmut. For \seg items, raters are presented with a single segment, whereas \doc and \docshuff items display a full document.}
    \label{fig:pearmut}
\end{figure}

\paragraph{Annotation protocol.}
\label{appx:annot_protocol}
We follow the ESA protocol~\citep{kocmi-etal-2024-error}, adapted for Korean. Annotators evaluate each translation along two axes: fluency and adequacy at the segment level, and discourse phenomena such as consistency, logical flow, formality, and coherence at the document level. The procedure has three steps: \circled{i} marking error spans, \circled{ii} labeling each span as \textit{Major} or \textit{Minor}, and \circled{iii} assigning an overall score from 0 to 100, anchored at four reference points (Nonsense, Broken, Middling, Perfect) as defined by Pearmut. \Cref{fig:pearmut} displays the segment- and document-level evaluation screens.

The annotation guideline was prepared in Korean and distributed to all annotators prior to the task, specifying the evaluation procedure, score anchors, error categories, and tool usage. Critically, annotators were not informed of the counterfactual design: \docshuff documents were presented under the same interface as \doc, with no indication that some documents combined outputs from multiple systems. This blinding is essential to the validity of our comparison, as awareness of the manipulation would induce task demand effects that obscure ESA's natural behavior on incoherent input.

\paragraph{Instructions on discourse phenomena.}
Blinding concerned the manipulation, not the phenomena. The guideline directed raters to discourse-level properties explicitly: its document-level section names consistency of terminology and entity forms, logical flow between sentences, formality and speech-level consistency, and overall coherence, with a worked Korean example of each. Raters were instructed to mark a span and lower the segment score when such a property is violated, exactly as they would for a fluency or adequacy error. The negative result is therefore not attributable to annotators having been left unaware that discourse phenomena were in scope.

We deliberately did not add a separate coherence-scoring item. Our claim concerns ESA \emph{as deployed}, whose premise is that showing the full document is sufficient for discourse errors to surface in the per-segment outputs. \docshuff is dense with such errors and, under that premise, should score lower. A dedicated coherence question would test a different instrument, and would in effect be one of the redesigned protocols we call for in \Cref{sec:conclusion} rather than a fix to the one under test.

\subsection{Automatic Metrics}
\label{appx:auto_metrics}

We use a suite of MT evaluation metrics covering surface, learned, and LLM-as-judge approaches at both segment and document granularities.

\subsection{Segment-level metrics}

\paragraph{\chrf~\citep{popovic-2015-chrf}.}
A character-level surface metric that computes the F-score of character $n$-grams between hypothesis and reference. \chrf was among the strongest-performing metrics at WMT25~\citep{lavie-etal-2025-findings}. We use \textsc{chrF++} (adding word $n$-grams up to order 2) for Korean.

\paragraph{\xcomet~\citep{guerreiro-etal-2024-xcomet}.}
A learned reference-based metric built on XLM-R XXL (11B parameters). It produces regression-style quality scores from (source, hypothesis, reference) triples and has been a top performer at recent WMT shared tasks. \xcometkiwi~\citep{rei-etal-2023-scaling} is its reference-free counterpart.

\paragraph{\metricx~\citep{juraska-etal-2024-metricx}.}
A learned metric based on mT5-XXL that predicts MQM-style quality scores. Unlike most metrics, \textsc{metricX} produces error scores (0 = perfect, 25 = worst), which we negate in our analyses for consistency with other metrics. \metricxqe is its reference-free variant.

\paragraph{\gemba~\citep{junczys-dowmunt-2025-gemba}.}
An LLM-as-judge metric. We use \mGpt as the underlying LLM, matching the configuration under which \gemba's performance was originally reported~\citep{junczys-dowmunt-2025-gemba}. The model is prompted with source-hypothesis pairs and asked to elicit a 0--100 quality rating following the \gemba prompt template, which extends the original \textsc{gemba}~\citep{kocmi-federmann-2023-gemba} with error-span elicitation prior to scoring.

\subsection{Document-level metrics}

\paragraph{\dbleu~\citep{liu-etal-2020-multilingual-denoising}.}
Corpus-level BLEU computed on full document hypotheses concatenated against full document references. \dbleu is sensitive to large-scale $n$-gram overlap rather than to discourse coherence.

\paragraph{\doccomet~\citep{vernikos-etal-2022-embarrassingly}.}
A document-aware extension of \textsc{comet} that incorporates surrounding sentences into the (source, hypothesis, reference) triple, allowing the model to use intra-document context.

\paragraph{\slide~\citep{raunak-etal-2023-evaluating}.}
A reference-free document-level metric (Sliding Document Evaluator) that scores translations through a sliding window over consecutive segments, aggregating per-window quality estimates from a base QE model (\xcometkiwi in our experiments).

\paragraph{\falcon~\citep{Kim_2025-falcon}.}
An LLM-as-judge metric designed for document-level MT evaluation. It prompts the LLM with full documents under nine discourse-related criteria and elicits both MQM-style error annotations and overall quality scores. We use \mGpt as the underlying LLM, matching the configuration under which \falcon's performance was originally reported.

\subsection{Implementation notes}

We exclude the \mRef system, since hypothesis-reference identity yields trivially perfect scores for reference-based metrics. Because segment-level metrics produce identical document-level scores for \seg and \doc by construction, we focus the metric analysis on the \doc vs.\ \docshuff comparison. Reference-based metrics (\chrf, \xcomet, \metricx, \dbleu, \doccomet) use the publicly available references included in the WMT dataset, while reference-free metrics (\xcometkiwi, \metricxqe, \slide, \gemba, \falcon) operate on (source, hypothesis) only. For all metrics except \metricx variants, higher scores indicate better quality; \metricx produces error scores (lower = better), which we negate in correlation and ranking analyses to maintain consistent direction across metrics.

\paragraph{Equivalence test (\Cref{h1}) details.}
We set the equivalence margin per metric scale: 3 points for 0--100 scales (\chrf, \dbleu), 0.05 for 0--1 scales (\xcomet variants), and 0.5 for the 25-point \metricx scale. All metrics support equivalence at TOST $p<0.001$.

\paragraph{Restriction to segment-level metrics for human$\bm{\leftrightarrow}$metric agreement.}
\label{appx:p5_metrics}
The human-metric agreement analysis is restricted to segment-level metrics. Document-level metrics produce one score per document, which under \doc can be attributed to the single system that translated the document but under \docshuff corresponds to documents combining segments from multiple systems. Distributing a single \docshuff document score across constituent systems requires modeling assumptions (such as proportional to segment count) that introduce analysis artifacts, so we restrict the agreement analysis to segment-level metrics, where each segment carries an unambiguous system label.

\paragraph{A note on token truncation.}
We apply \xcomet and \metricx variants at the document level for completeness. However, both models have maximum input lengths shorter than typical document lengths in our data. \xcomet accepts up to 512 tokens, while \metricx accepts up to 1{,}536 tokens. Documents are therefore truncated under both metrics, but the effect is most severe for 
\xcomet-doc which yields very low scores ($\sim 0.16$ in both conditions), compared with the typical \xcomet range of $0.5$--$0.9$ on segment-level inputs. \metricx-doc variants are less affected, owing to the longer input window. We include all document-level results for completeness but interpret \xcomet-doc as reflecting truncation rather than genuine document-level evaluation.

\newpage
\section{Statistical Methodology}
\label{appx:evaluation}

\subsection{Equivalence Testing (TOST)}
The Two One-Sided Tests (TOST) procedure formally tests the null hypothesis $H_0: |\mu_1 - \mu_2| \geq \epsilon$ against the alternative $H_1: |\mu_1 - \mu_2| < \epsilon$, where $\epsilon$ is the smallest difference of practical interest~\citep{Fay2010}. A significant TOST result supports the conclusion that the two conditions are practically equivalent within $\pm\epsilon$. Unlike a non-significant conventional test, which cannot establish equivalence, TOST shifts the burden of proof. Equivalence must be demonstrated.

\paragraph{Equivalence margin.}
We set $\epsilon = 3$ points on the 100-point ESA scale on two grounds. First, it corresponds to approximately $0.2$ standard deviations of the empirical score distribution ($\sigma \approx 13$--$15$). Second, it lies below the typical magnitude of IAA in ESA evaluation. Equivalence at $\epsilon = 3$ implies equivalence at any larger margin. We additionally report results at $\epsilon \in \{5, 7\}$ in \Cref{tab:tost_full} for transparency.

\paragraph{Paired and independent TOST.}
The main results report \emph{paired TOST}, which matches observations on the same segment evaluated under both conditions, controlling for between-segment variance. We additionally report \emph{independent-sample TOST} as a robustness check (\Cref{tab:tost_full}). It treats observations as independent samples and yields consistent conclusions across all comparisons.

\subsection{Ranking Analysis [\Cref{h2}]}
\label{appx:ranking}

System rankings are induced from segment-level mean scores aggregated by system, over the $n = 9$ MT systems, excluding the human reference \mRef. This yields $\tbinom{9}{2} = 36$ pairwise comparisons. (\Cref{tab:h2_combined} additionally displays \mRef for human ESA, where it is scored like any other system.) We compute Kendall's $\tau$ between ranking pairs $(r_a, r_b)$ for conditions $a, b \in \{\seg, \doc, \docshuff\}$, preferring $\tau$ over Spearman's $\rho$ given the small number of systems and of tied ranks. The threshold $\tau_{\min} = 0.667$ corresponds to moderate ranking agreement, and values below it indicate re-ordering across conditions.

\paragraph{Bootstrap confidence intervals.}
We report $95\%$ bootstrap confidence intervals computed by resampling source documents with replacement ($B = 10{,}000$ resamples). Resampling is at the document level because annotations are nested within the 23 source documents and the segment-level resampling would understate the variance. For each resample we recompute system-level mean scores, induce rankings, and compute $\tau$, taking the $2.5$th and $97.5$th percentiles of the resulting distribution.

For human ESA, $\tau(\doc, \docshuff)$ has a bootstrap mean of $0.737$ ($\mathrm{SD} = 0.097$) and a $95\%$ CI of $[0.539, 0.911]$, and $22.5\%$ of resamples fall below $\tau_{\min}$. Expressed in discordant pairs, the point estimate of $0.778$ corresponds to $5$ of $45$ pairs ordered differently under the two conditions. The pairs that reverse are not arbitrary: they are the pairs whose mean scores are closest under \doc, where the difference between systems is smaller than the sampling variability of the means themselves (\mGpt and \mWenyiil differ by $0.05$ ESA points, \mGemini and \mRef by $0.19$). Exact ties among such pairs arise in $1.1\%$ of resamples and are handled by the $\tau_b$ correction. $\tau$ assigns a strict order to these pairs regardless, so its value is largely determined by which documents enter a given resample. We therefore treat the clustering analysis in \Cref{tab:h2_combined} as the primary evidence for \Cref{h2}.

\paragraph{Statistical clustering.}
Following the WMT methodology~\citep{kocmi-etal-2025-findings}, we group systems into statistically equivalent clusters via pairwise bootstrap tests ($B = 1{,}000$ resamples, $\alpha = 0.05$; the smaller $B$ follows WMT and suffices for a significance decision, whereas the interval estimates above require more resamples in the tails). Two systems are placed in the same cluster when their mean score difference is not significantly different from zero, so that pairs of the kind described above are left unordered rather than assigned an arbitrary rank. This yields condition-specific cluster assignments that are stable under small score perturbations.

\subsection{Error Annotation Analysis [\Cref{h3}]}
\label{appx:errors}

We compute error counts per 1{,}000 target characters to control for varying segment length. Paired TOST compares $\eta_{\mathrm{\small MIX}}$ against $\eta_{\mathrm{\small DOC}}$ on a per-segment basis with margin $\epsilon_{\mathrm{err}} = 0.5$ errors per annotation (\Cref{h3}). We additionally report the proportion of annotations marked as error-free (score $= 100$) and the distribution of error severity (Major/Minor) across conditions.

\paragraph{Positional analysis of error spans.}
For signal \circled{3} we locate each span by the character offset of its midpoint within its segment, normalized to $[0, 1]$ by segment length. A span counts as \emph{boundary-localized} if that position falls in $[0, 0.2)$ or $(0.8, 1]$, that is, within the first or last fifth of the segment. Under a null of no positional structure the expected rate is $40\%$, and the observed rates near $50\%$ reflect the tendency of annotated spans to include sentence-initial and sentence-final material in all conditions alike, which is why the between-condition contrast, not the absolute level, carries the inference. Confidence intervals are bootstrap intervals over segments ($B = 10{,}000$).

Three robustness variants leave the result unchanged. (i) Excluding zero-width omission markers, which carry no span extent and are placed by convention at a segment edge. (ii) Restricting to document-interior segments, since the first and last segment of a document have only one neighboring seam. (iii) Restricting to \textit{Major} spans, on the reasoning that a register or terminology break should be marked as major if it is marked at all. In every variant the \docshuff${}-{}$\doc difference remains within $\pm 1.5$ points with a confidence interval covering zero.

\subsection{Behavioral Measure}
\label{appx:behavioral}

\paragraph{IAA.}
We compute Krippendorff's $\alpha$ on the interval scale at two units of analysis. At the document level, the unit is $\langle\texttt{doc\_id}, \texttt{system}\rangle$, with per-annotator scores averaged across the constituent segments. At the segment level, the unit is $\langle\texttt{doc\_id}, \texttt{seg\_id}, \texttt{system}\rangle$.

\paragraph{Annotation time.}
Per-segment annotation time is computed as the cumulative time between consecutive annotation actions within a task, excluding gaps longer than 120 seconds (treated as inactive periods), divided by the number of segments in the task. This definition isolates active evaluation effort from incidental pauses.

\begin{table*}[t]
\centering
\small
\begin{tabular}{llcccccc}
\toprule
& & & & & \multicolumn{3}{c}{\textbf{TOST $p$-value}} \\
\cmidrule(lr){6-8}
\textbf{Comparison} & \textbf{Design} & \textbf{n} & \textbf{Mean diff.} & \textbf{90\% CI} & $\varepsilon{=}3$ & $\varepsilon{=}5$ & $\varepsilon{=}7$ \\
\midrule
\rowcolor{yellow!12}
\doc\ vs.~\docshuff\ & independent & 5920/5920 & -0.82 & [-1.25, -0.40] & $<$0.001 & $<$0.001 & $<$0.001 \\
\rowcolor{yellow!12}
 & paired & 308 & -1.00 & [-2.05, +0.04] & $<$0.001 & $<$0.001 & $<$0.001 \\
\midrule
\doc\ vs.~\seg\ & independent & 5920/5920 & +0.55 & [+0.11, +0.99] & $<$0.001 & $<$0.001 & $<$0.001 \\
 & paired & 364 & +0.07 & [-0.83, +0.97] & $<$0.001 & $<$0.001 & $<$0.001 \\
\midrule
\seg\ vs.~\docshuff\ & independent & 5920/5920 & -1.37 & [-1.79, -0.95] & $<$0.001 & $<$0.001 & $<$0.001 \\
 & paired & 1249 & -1.09 & [-1.65, -0.52] & $<$0.001 & $<$0.001 & $<$0.001 \\
\bottomrule
\end{tabular}
\caption{TOST equivalence test results across all pairwise condition comparisons, designs (independent vs.\ paired), and equivalence margins $\varepsilon \in \{3, 5, 7\}$ points on the 100-point ESA scale. For independent designs, $n$ is reported as $n_1/n_2$; for paired designs, $n$ is the number of matched pairs. All eighteen cells support equivalence at $p<0.001$, demonstrating robustness to comparison choice, study design, and equivalence margin.}
\label{tab:tost_full}
\end{table*} 
\newpage
\section{Detailed Results}

\subsection{Score Equivalence [\Cref{h1}]}
\label{appx:p1_further}

\paragraph{TOST across pairwise comparison and metrics.}
\label{appx:tost}

\Cref{tab:tost_full} reports TOST results for all three pairwise comparisons (\doc vs.\ \docshuff, \doc vs.\ \seg, \seg vs.\ \docshuff) under both independent and paired designs. Independent and paired analyses yield the same qualitative conclusion for every comparison, with paired estimates consistently closer to zero. For \doc vs.\ \seg, the mean difference shifts from $+0.55$ under the independent design to $+0.07$ under the paired design. For \seg vs.\ \docshuff, it shifts from $-1.37$ to $-1.09$. Part of the apparent between-condition variance is therefore attributable to inter-annotator differences rather than to genuine condition effects.

\paragraph{Robustness to margin choice.}
We report results at equivalence margins $\epsilon \in \{3, 5, 7\}$ in \Cref{tab:tost_full}. All eighteen cells (three comparisons $\times$ two designs $\times$ three margins) yield $p < 0.001$, with mean differences ranging from $-1.37$ to $+0.55$. The equivalence conclusion holds across comparison choice, study design, and margin value within the range tested.

\subsection{IAA}
\label{appx:iaa}

\paragraph{Document- vs.\ segment-level agreement.}
Krippendorff's $\alpha$ depends on the unit of analysis. \Cref{tab:iaa} reports both units (see \Cref{appx:behavioral} for definitions). At the document level, \doc leads with $\alpha = 0.186$. At the segment level, the order reverses, with \docshuff leading ($\alpha = 0.225$) and \doc lowest ($\alpha = 0.155$). The pattern suggests that under \doc, annotators incorporate document context, producing diverse per-segment judgments while converging at the document level on shared overall impressions. Under \docshuff, the absence of coherent document context returns judgments to segment-only assessment, yielding tighter segment-level agreement.

\begin{table}[t]
\centering
\small
\begin{tabular}{@{}lcc@{}}
\toprule
& \multicolumn{2}{c}{Krippendorff's $\alpha$} \\
\cmidrule(l){2-3}
Condition & Per document & Per segment \\
\midrule
\seg       & 0.149 & 0.172 \\
\doc       & \textbf{0.186} & 0.155 \\
\docshuff  & 0.128 & \textbf{0.225} \\
\bottomrule
\end{tabular}
\caption{Inter-annotator agreement, computed at two units. \textit{Per document} treats $\langle$\texttt{doc\_id}, \texttt{system}$\rangle$ as the item, averaging each annotator's segment scores within it; \textit{per segment} treats $\langle$\texttt{doc\_id}, \texttt{seg\_id}, \texttt{system}$\rangle$ as the item. The orderings invert between the two: \doc agrees most per document but least per segment. All values fall below the conventional threshold for acceptable reliability ($\alpha \geq 0.667$), consistent with known variability in scalar MT evaluation~\citep{graham-etal-2013-continuous,kocmi-etal-2025-findings}.}
\label{tab:iaa}
\end{table}

\paragraph{A note on \docshuff's document-level agreement.}
For \docshuff, the document-level item $\langle\texttt{doc\_id}, \texttt{system}\rangle$ does not correspond to a single physical document. Each \docshuff document is a unique recombination of segments from multiple systems, and the same $\langle\texttt{doc\_id}, \texttt{system}\rangle$ pair may refer to different segment sets across annotators. The lower document-level $\alpha$ for \docshuff is therefore largely a structural artifact rather than a behavioral signal. Our analysis emphasizes the segment-level $\alpha$ comparison, where item identity is unambiguous.

\paragraph{Absolute reliability remains low.}
All values fall below the conventional threshold for acceptable reliability ($\alpha \geq 0.667$), in line with known variability in scalar MT evaluation. This does not compromise our primary findings. The analyses operate on aggregated scores derived from large per-condition samples ($n = 5{,}920$ each), where annotator-level noise is averaged out, and within-annotator paired analyses in \Cref{sec:analysis_p1} explicitly control for inter-annotator variance.

\paragraph{Per-annotator normalization.}
Annotator-wise $z$-score normalization shifts the absolute values but preserves the ordering across conditions. After normalization, document-level $\alpha$ for \doc rises from $0.186$ to $0.23$, while \seg and \docshuff remain around $0.15$ and $0.13$ respectively. The document-level result is therefore not an artifact of annotator-specific scoring styles.

\begin{table}[t]
\centering
\small
\resizebox{\linewidth}{!}{
\begin{tabular}{llccc}
\toprule
Metric & Condition & Kendall's $\tau$ & $p$ & Pearson's $r$ \\
\midrule
\textit{Surface} \\
\quad \chrf & \doc\      & $0.667$ & $0.013$ & $0.815$ \\
           & \docshuff\ & $0.889$ & $<0.001$ & $0.946$ \\
\quad \dbleu & \doc\      & $0.667$ & $0.013$ & $0.831$ \\
           & \docshuff\ & $0.889$ & $<0.001$ & $0.950$ \\
\midrule
\textit{Learned ref-based} \\
\quad \xcomet  & \doc\      & $0.500$ & $0.075$ & $0.642$ \\
              & \docshuff\ & $0.611$ & $0.025$ & $0.803$ \\
\quad \metricx & \doc\      & $0.556$ & $0.045$ & $0.604$ \\
              & \docshuff\ & $0.667$ & $0.013$ & $0.844$ \\
\midrule
\textit{Reference-free (QE)} \\
\quad \xcometkiwi & \doc\      & $0.222$ & $0.477$ & $0.252$ \\
                  & \docshuff\ & $0.333$ & $0.260$ & $0.505$ \\
\quad \metricxqe  & \doc\      & $0.278$ & $0.359$ & $0.278$ \\
                  & \docshuff\ & $0.389$ & $0.180$ & $0.588$ \\
\midrule
\textit{LLM-as-judge} \\
\quad \gemba & \doc\      & $0.504$ & $0.075$ & $0.652$ \\
             & \docshuff\ & $0.722$ & $0.007$ & $0.872$ \\

\bottomrule
\end{tabular}
}
\caption{System-level human-metric correlations by condition ($n=9$ systems). Kendall's $\tau$ values for QE metrics (\xcometkiwi, \metricxqe) are not statistically significant at $p<0.05$, reflecting the small system pool and narrow score distributions of these metrics. We focus on the relative pattern across conditions in our analysis.}
\label{tab:human_metric_full}
\end{table}
\begin{table}[t]
\centering
\small
\resizebox{\linewidth}{!}{
\begin{tabular}{llcccc}
\toprule
& Scale & Range & SD & Min & Max \\
\midrule
Human evaluation & 0--100 & $12.28$ & $4.09$ & $78.60$ & $90.88$ \\
\midrule
\multicolumn{6}{l}{\textit{Surface metrics}} \\
\quad \chrf  & 0--100 & $9.81$  & $3.08$ & $23.37$ & $33.18$ \\
\quad \dbleu & 0--100 & $11.66$ & $3.64$ & $15.52$ & $27.18$ \\
\midrule
\multicolumn{6}{l}{\textit{Learned reference-based}} \\
\quad \xcomet      & 0--1  & $0.076$ & $0.032$ & $0.716$ & $0.792$ \\
\quad \metricx$^*$ & 0--25 & $1.030$ & $0.382$ & $3.826$ & $4.856$ \\
\midrule
\multicolumn{6}{l}{\textit{Reference-free (QE)}} \\
\quad \xcometkiwi    & 0--1  & $0.056$ & $0.016$ & $0.806$ & $0.861$ \\
\quad \metricxqe$^*$ & 0--25 & $0.839$ & $0.268$ & $3.602$ & $4.441$ \\
\midrule
\multicolumn{6}{l}{\textit{LLM-as-judge}} \\
\quad \gemba & 0--100 & $8.43$ & $2.71$ & $79.52$ & $87.95$ \\
\bottomrule
\end{tabular}
}
\caption{System-level mean score distributions across nine systems under each metric. Range is computed as max $-$ min of system means. Reference-free QE metrics produce the narrowest output ranges ($\sim 3$--$6\%$ of metric scale), making ranking-based human agreement noise-sensitive. Surface metrics span ranges comparable to human evaluation. $^*$\textsc{metricX} variants are error scores (lower = better); ranges are reported in the metric's native scale.}
\label{tab:metric_spread}
\end{table}
 
\subsection{Human--Metric Agreement}
\label{appx:metric_result}
We provide detailed per-metric results. Kendall's $\tau$ and Pearson's $r$ at $n = 9$ are therefore noise-sensitive, and we emphasize relative patterns across conditions rather than absolute correlation values.

\paragraph{Per-metric, per-condition correlations.}
\Cref{tab:human_metric_full} reports detailed human-metric system correlations for both \doc and \docshuff conditions, complementing \Cref{sec:mechanism}. For surface metrics (\chrf, \dbleu), agreement shifts from $\tau = 0.67$ under \doc to $\tau = 0.89$ under \docshuff. The same direction holds for learned metrics at lower absolute values. \xcomet shifts from $0.50$ to $0.61$, and \metricx from $0.56$ to $0.67$. Reference-free QE variants show the same upward shift from a lower baseline, from $\tau = 0.22$--$0.28$ under \doc to $0.33$--$0.39$ under \docshuff. The LLM-as-judge metric \gemba follows the same pattern, moving from $\tau = 0.50$ under \doc to $0.72$ under \docshuff—matching surface metrics in the magnitude of its \docshuff shift while sitting between learned reference-based and surface metrics in absolute agreement.

\paragraph{On the narrow output range of QE metrics.}
\Cref{tab:metric_spread} reports system-level score distributions. Surface metrics (\chrf, \dbleu) span $\sim 10\%$ of their scale, comparable to human evaluation ($\sim 12\%$), and \gemba spans a similar $\sim 8\%$. Learned reference-based metrics are narrower ($\sim 4$--$8\%$), and reference-free QE metrics are narrowest of all: \xcometkiwi at $\sim 5.6\%$ (range $= 0.056$) and \metricxqe at $\sim 3.4\%$ (range $= 0.84$ on a 25-point scale). With distributions this narrow, ranking-based agreement becomes noise-sensitive, and small score fluctuations can flip ranks—plausibly contributing to QE metrics' lower absolute correlation with humans.

\end{document}